\documentclass[a4paper,fleqn]{cas-dc}

\usepackage{setspace}
\usepackage{amsmath,amssymb,amsfonts}

\usepackage{bbm} 
\usepackage{bm}
\usepackage{algorithmic}
\usepackage{algorithm}
\usepackage{tikz}
\usepackage{comment}
\usepackage{color}
\usepackage{overpic}
\usepackage{graphicx}
\usepackage{pifont}
\usepackage{pict2e} 
\usepackage{booktabs}
\usepackage{multirow}
\usepackage{multicol}
\usepackage{xcolor,colortbl}
\usepackage{nicematrix}

\definecolor{Gray}{gray}{0.95}
\definecolor{darkgreen}{rgb}{0,0.5,0}
\definecolor{LightCyan}{rgb}{0.88,1,1}
\definecolor{LightRed}{rgb}{1,0.9,0.9}

\newcommand{\myPara}[1]{\par\vspace{.0in}\textbf{#1.}}

\usepackage[capitalize,noabbrev]{cleveref}
\crefname{section}{Sec.}{Secs.}
\crefname{figure}{Fig.}{Figs.}
\crefname{table}{Tab.}{Tabs.}
\crefname{equation}{Eq.}{Eqs.}
\crefname{algorithm}{Alg.}{Algs.}

\definecolor{color1}{RGB}{237,125,49}
\definecolor{color3}{RGB}{0,112,192}
\definecolor{color2}{RGB}{112,173,71}
\usepackage{etoolbox}
\usepackage{listofitems}
\usepackage{wheelchart}
\usetikzlibrary{decorations.text}
\usetikzlibrary{decorations.markings}
\usepackage{makecell}
\usepackage{wrapfig}

\usepackage{forest}
\usetikzlibrary{trees,positioning,shapes,shadows,arrows.meta}

\makeatletter
\renewcommand\subsubsection{\@startsection{subsubsection}{4}{1\parindent}{0ex plus 0.1ex minus 0.1ex}{0ex}{\normalfont\normalsize\bfseries\itshape}}\renewcommand\paragraph{\@startsection{paragraph}{4}{2\parindent}{0ex plus 0.1ex minus 0.1ex}{0ex}{\normalfont\normalsize\bfseries\itshape}}\makeatother

\usetikzlibrary{trees}
\definecolor{hidden-draw}{rgb}{0.5, 0.5, 0.5}
\definecolor{harvestgold}{rgb}{0.85, 0.57, 0.0}
\definecolor{cyan}{rgb}{0.0, 1.0, 1.0}
\definecolor{lightcoral}{rgb}{0.94, 0.5, 0.5}
\definecolor{SlateBlue}{rgb}{0.416, 0.353, 0.804}
\definecolor{LightGreen}{rgb}{0.564, 0.933, 0.564}
\definecolor{Turquoise}{rgb}{0.251, 0.878, 0.816}
\definecolor{BlueGreen}{rgb}{0.643, 0.859, 0.871}
\definecolor{BlueViolet}{rgb}{0.54, 0.17, 0.89}
\definecolor{RAG}{rgb}{0.639, 0.835, 0}
\definecolor{0}{rgb}{0.92, 0.3, 0.26}
\definecolor{1}{rgb}{0.961, 0.851, 0.471}
\definecolor{1_1}{rgb}{1.00, 0.933, 0.698}
\definecolor{2}{rgb}{0.522, 0.784, 0.953}
\definecolor{2_1}{rgb}{0.722, 0.592, 0.871}
\definecolor{2_2_2}{rgb}{0.561, 0.659, 0.843}
\definecolor{3}{rgb}{0.56, 0.93, 0.56}
\definecolor{3_1_1}{rgb}{0.808, 0.996, 0.808}
\definecolor{4}{rgb}{0.988, 0.541, 0.565}
\definecolor{4_1}{rgb}{0.996, 0.796, 0.808}
\definecolor{4_1_1}{rgb}{0.957, 0.847, 0.843}
\definecolor{bounding}{rgb}{0.855, 0.392, 0.357}
\definecolor{6}{rgb}{0.690, 0.612, 0.522}

\definecolor{golden}{RGB}{255, 215, 0}
\definecolor{golden}{RGB}{0, 0, 0}

\definecolor{backbone}{RGB}{255,192,0}
\definecolor{neck}{RGB}{112,173,71}
\definecolor{head}{RGB}{237,125,49}
\definecolor{score}{RGB}{68,72,212}
\definecolor{grade}{RGB}{92,201,207}
\definecolor{rank}{RGB}{0,176,207}

\usepackage{fontawesome}
\usepackage{nicematrix}
\usepackage{calc}

\usepackage{orcidlink}
\usepackage{ragged2e}

\usepackage{tikz}
\usetikzlibrary{calc,positioning}

\newlength{\myMheight}
\usepackage{bbding,pifont}

\usepackage[misc]{ifsym}

\usepackage{times}

\usepackage{etoolbox}
\makeatletter
\patchcmd{\thebibliography}
  {\settowidth}
  {\setlength{\itemsep}{0pt plus 0.3ex}
   \setlength{\parskip}{0pt}
   \settowidth}
  {}{}
\makeatother

\graphicspath{{./figs/}{./figs/imgs/}}
\usepackage{overpic}
\usepackage{subfigure}

\usepackage{subcaption}
\usepackage{xstring}   

\colorlet{highlightcolor}{red!70!black}
\newif\ifMShighlight
\MShighlighttrue      
\MShighlightfalse     

\newif\ifMSinputmode
\MSinputmodefalse

\usepackage[pagewise]{lineno}
\linenumbers

\ifMShighlight
    \else
    
\fi

\newcommand{\MSsection}[2]{\ifMSinputmode
    \section{\textcolor{highlightcolor}{#2}}\else
    \section{#2}\fi
  \label{#1}}

\newcommand{\MSfigcaption}[3]{\ifMSinputmode
    \caption{\textcolor{highlightcolor}{#2}}\else
    \caption{#2}\fi
  \label{#1}#3}

\newcommand{\MStabcaption}[2]{\ifMSinputmode
    \caption{\textcolor{highlightcolor}{#2}}\else
    \caption{#2}\fi
  \label{#1}}

\makeatletter
\let\MS@originput\input
\renewcommand{\input}[1]{\IfBeginWith{#1}{pr/r2/}{
    \ifMShighlight
      \begingroup
        \MSinputmodetrue        \linelabel{start:#1}\color{highlightcolor}\arrayrulecolor{highlightcolor}\MS@originput{#1}\arrayrulecolor{black}\linelabel{end:#1}\endgroup
      \ignorespaces
    \else 
      \MS@originput{#1}\ignorespaces
    \fi
  }{\MS@originput{#1}\ignorespaces
  }}
\makeatother

\hypersetup{%
  linkcolor={blue},
  urlcolor={magenta},
  citecolor={blue}
}

\def\tsc#1{\csdef{#1}{\textsc{\lowercase{#1}}\xspace}}
\tsc{WGM}
\tsc{QE}
\tsc{EP}
\tsc{PMS}
\tsc{BEC}
\tsc{DE}

\makeatletter
\renewcommand{\vspace}{\@ifstar\@gobble\@gobble}
\makeatother
\usepackage{fancyvrb}
\begin{document}
\let\WriteBookmarks\relax
\def\floatpagepagefraction{1}
\def\textpagefraction{.001}

\shorttitle{NeuroPath: Brain-Inspired Dual-Pathway Graph Convolutional Networks for Skeleton-Based Action Recognition}

\shortauthors{K. Zhou et~al.}

\title [mode = title]{NeuroPath: Brain-Inspired Dual-Pathway Graph Convolutional Networks for Skeleton-Based Action Recognition}

\author[1,2]{Kanglei Zhou}
\ead{zhoukanglei@tsinghua.edu.cn}
\credit{Writing -- original draft, Formal analysis}

\author[2]{Ruizhi Cai}
\ead{craaaaazy@buaa.edu.cn}
\credit{Formal analysis, Data curation, Conceptualization}

\author[3]{Hubert P. H. Shum}
\ead{hubert.shum@durham.ac.uk}
\credit{Writing -- review \& editing}

\author[3]{Frederick W. B. Li}
\ead{frederick.li@durham.ac.uk}
\credit{Writing -- review \& editing}

\author[2,4]{Xiaohui Liang}[orcid=0000-0001-6351-2538]
\cormark[1]
\ead{liang_xiaohui@buaa.edu.cn}
\credit{Writing -- review \& editing, Supervision}

\cortext[cor1]{Corresponding author}

\affiliation[1]{
    organization={Department of Psychological and Cognitive Sciences, Tsinghua University},
    addressline={No. 30 Shuangqing Road, Haidian District},
    city={Beijing},
    postcode={100084},
    country={China}
}

\affiliation[2]{
    organization={State Key Laboratory of Virtual Reality Technology and Systems, Beihang University},
    addressline={No. 37 Xueyuan Road, Haidian District},
    city={Beijing},
    postcode={100191},
    country={China}
}

\affiliation[3]{
    organization={Department of Computer Science, Durham University},
    addressline={Stockton Rd},
    city={Durham},
    postcode={DH1 3LE},
    country={United Kingdom}
}

\affiliation[4]{
    organization={Zhongguancun Laboratory},
    city={Beijing},
    country={China}
}


\begin{abstract}
    Skeleton-based action recognition aims to recognize human actions from sequences of human joint coordinates.
    Most existing Spatial-Temporal Graph Convolutional Networks (STGCNs) have achieved promising results by modeling skeletal structures with implicit spatial–temporal representations.
    However, our empirical study reveals a clear performance imbalance across different skeletal modalities, indicating that implicitly coupling spatial and temporal information limits the full exploitation of complementary structural and motion cues.
    Inspired by the ventral and dorsal pathways in human perception, we propose Dual-Pathway Graph Convolutional Networks (NeuroPath), which adopt a dual-pathway architecture for separate yet collaborative modeling of spatial and temporal information.
    Specifically, transformation units first convert the input into pathway-specific skeletal representations, allowing each pathway to focus on complementary aspects of human motion.
    To further capture coordinated joint behaviors and their interrelationships, we introduce a group graph convolution block that dynamically identifies key body parts and models their spatial-temporal dependencies.
    In addition, inter-pathway dynamic fusion modules integrate complementary inter-modal information across pathways, facilitating higher-level semantic interpretation of actions.
    Extensive experiments on Kinetics Skeleton 400, NTU RGB+D 60, and NTU RGB+D 120 demonstrate consistent performance improvements, validating the effectiveness of dual-pathway spatial-temporal modeling for skeleton-based action recognition.
\end{abstract}




\thispagestyle{empty}










\maketitle

\section{Introduction}
\label{sec:intro}
Human action recognition aims to recognize human behaviors~\cite{zhou25phi,zhu2023motionbert}, and is fundamental to applications such as security surveillance, autonomous driving, and human--computer interaction \cite{zhao2022place,zhao2021classifying}.
Skeletal data provide a compact and structured representation of human motion, and are more robust to background clutter and illumination variations than raw video data~\cite{zhou2024comprehensive}.
These advantages have led to the widespread adoption of skeleton-based action recognition~\cite{li2021memory}.
However, accurately modeling skeletal data remains challenging, as human actions involve complex spatial dependencies among joints and long-term temporal dynamics that must be jointly captured.

\begin{figure}
    \centering
    \rm
    \begin{overpic}[width=\linewidth, clip, trim=0 5 0 5]{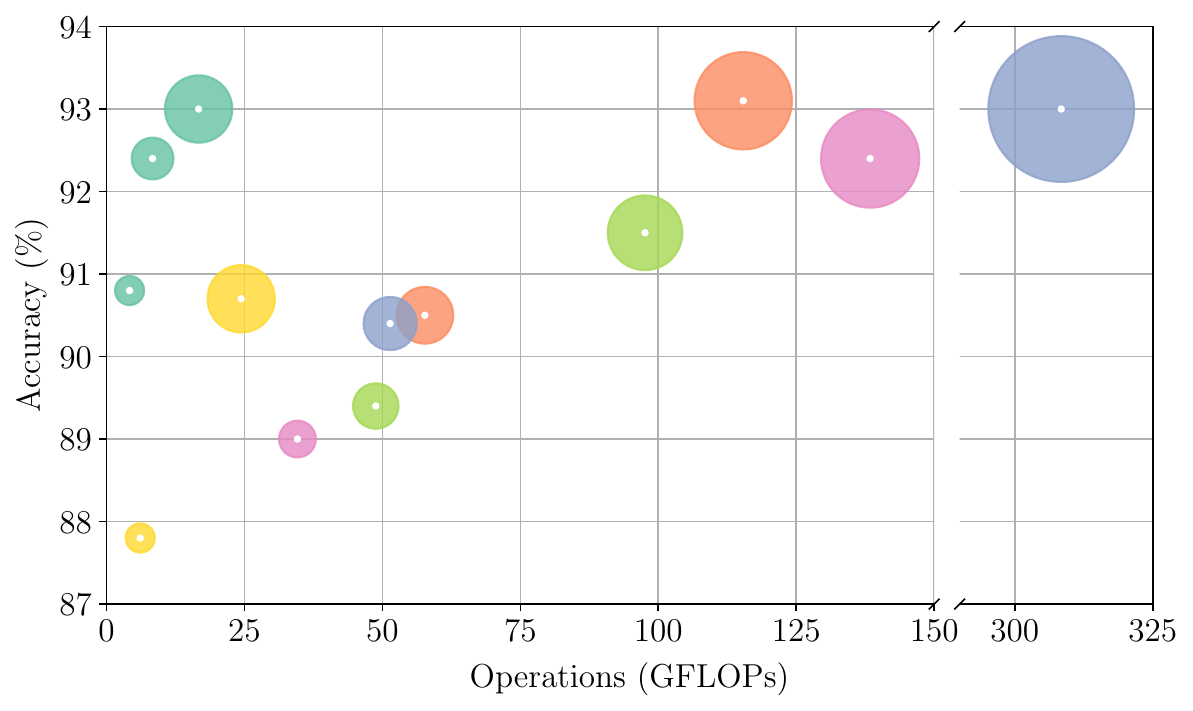}
        \put(10,37){\textcolor{red}{\scriptsize Ours (Js)}}
        \put(15,43.5){\textcolor{red}{\scriptsize Ours (2s)}}
        \put(20,49){\textcolor{red}{\scriptsize Ours (4s)}}
        \put(39,31){\scriptsize Js HetGCN \cite{gao2023learning}}
        \put(38,49){\scriptsize 2s HetGCN \cite{gao2023learning}}
        \put(12,28){\scriptsize Js InfoGCN \cite{chi2022infogcn}}
        \put(76,18){\scriptsize 6s InfoGCN \cite{chi2022infogcn}}
        \multiput(82,21)(0.5, 2){12}{\linethickness{0.1mm}\color{blue}\line(1, 4){0.2}}
        \put(27,18){\scriptsize Js CTR-GCN \cite{chen2021channel}}
        \put(46,43.5){\scriptsize 4s CTR-GCN \cite{chen2021channel}}
        \put(35,24){\scriptsize Js MS-G3D \cite{liu2020disentangling}}
        \put(58,38){\scriptsize 2s MS-G3D \cite{liu2020disentangling}}
        \put(14,13){\scriptsize Js Shift-GCN \cite{cheng2020skeleton}}
        \put(23,38){\scriptsize 4s Shift-GCN \cite{cheng2020skeleton}}
        \multiput(30,37)(-2, -0.5){4}{\linethickness{0.1mm}\color{blue}\line(-4,-1){0.6}}
    \end{overpic}
    \caption{
        \textbf{The bubble plot of model computational performance in the X-Sub setting on the NTU RGB+D 60 dataset.} The horizontal axis represents computational operations (GFLOPs), the vertical axis represents accuracy (\%), and the size of the bubbles represents the number of parameters (M). Different bubble colors represent different kinds of models.
    }
    \label{fig-bubble}
\end{figure}
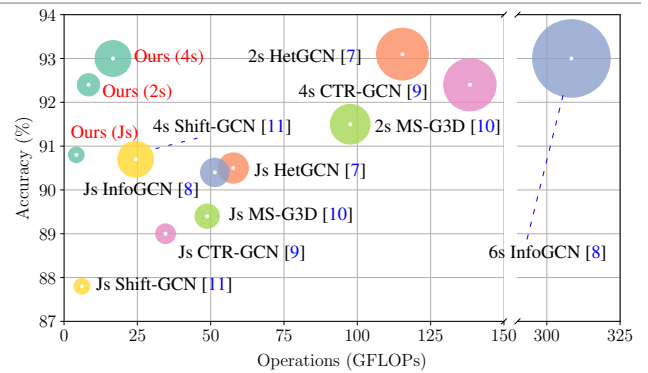

Early deep learning methods \cite{ren2020survey} using RNNs or CNNs have shown clear performance advantages over handcrafted approaches.
However, these methods typically represent skeleton sequences as vectors or pseudo-graphs, without explicitly modeling the structural constraints among body joints.
To better capture such structural information, Graph Convolutional Networks (GCNs) have attracted increasing attention due to their ability to naturally encode joint dependencies \cite{zhou2023hierarchical}.
The vanilla GCN~\cite{kipf2016semi}, however, focuses primarily on spatial relationships and neglects the temporal dynamics that are critical for modeling sequential human actions.
To address this limitation, Spatial-Temporal GCNs (STGCNs)~\cite{yan2018spatial,shi2019skeleton} have been proposed to jointly model spatial and temporal information, and have become a dominant paradigm for skeleton-based action recognition.
Despite these advances, effectively capturing the interplay between spatial structure and temporal dynamics in skeletal data remains a non-trivial challenge.

A closer examination of existing STGCNs reveals that this challenge is typically addressed by learning different skeletal modalities using separate streams, followed by simple score-level ensemble fusion~\cite{shi2019two,liu2020disentangling}.
Here, a \textbf{skeletal modality} denotes a specific representation derived from skeletal data, including joint positions, bone vectors, joint motion, and bone motion, which predominantly emphasize either spatial structure or temporal dynamics.
While such multi-stream ensemble strategies can improve overall performance, they implicitly assume that different modalities contribute independently.
However, our empirical analysis exposes a \textbf{clear performance imbalance}, where joint- and bone-based streams consistently outperform motion-based streams, indicating that motion-only representations lack sufficient structural support.
At the same time, the results also reveal strong \textbf{modality complementarity}: although motion-based streams perform poorly in isolation, integrating structural and motion modalities leads to substantial performance gains, whereas fusing motion-only modalities yields the lowest accuracy.
These findings motivate us to explicitly model the imbalanced yet complementary spatial and temporal information within a unified representation, rather than relying on loosely coupled ensemble strategies.

Motivated by the above observations and analysis, we revisit spatial–temporal modeling in skeleton-based action recognition from a modality-aware perspective.
We propose NeuroPath, a brain-inspired dual-pathway network that explicitly disentangles and collaboratively models spatial structure and temporal dynamics.
By separating yet coordinating spatial-aware and temporal-aware representations within a unified framework, NeuroPath aims to overcome the limitations of loosely coupled multi-branch designs and enable more effective exploitation of complementary skeletal modalities.

To mitigate inadequate inter-modal fusion, we design a modality transformation unit for each pathway, which decomposes the input skeletal data into complementary representations.
Specifically, one pathway is guided to focus on spatial-aware information, while the other emphasizes temporal-aware information, resembling the functional segregation of visual stimuli in the human brain.
Based on these pathway-specific representations, we further introduce a dynamic fusion module to explicitly model inter-modal dependencies at multiple stages of the network, enabling effective information exchange across pathways \cite{van2015interactions}.

To improve spatial–temporal cohesion within each pathway, we propose a spatial–temporal group graph convolution block.
First, the group aggregation module incorporates temporal cues into the spatial domain, facilitating the identification of key body parts.
Then, the adaptive structural aggregation module effectively infers the inter-part dependencies while avoiding redundancy.
Finally, a spatial-temporal attention module prevents discriminative information loss.

Extensive experiments on Kinetics Skeleton 400, NTU RGB+D 60, and NTU RGB+D 120 demonstrate consistent performance improvements, while achieving notable computational efficiency (see \cref{fig-bubble}).
The code is publicly available at \url{https://github.com/ZhouKanglei/NeuroPath}.
The main contributions of this work are:
\begin{itemize}
    \item We empirically uncover and analyze an overlooked yet critical phenomenon in skeleton-based action recognition, showing that different skeletal modalities exhibit intrinsic performance imbalance while remaining strongly complementary.
    \item Motivated by these observations, we introduce a brain-inspired dual-pathway framework that explicitly disentangles spatial-aware and temporal-aware representations while enabling their coordinated interaction within a single network.
    \item We provide an efficient and practical implementation of this framework, achieving consistent performance improvements with favorable computational efficiency on multiple skeleton-based action recognition benchmarks.
\end{itemize}

\section{Related Work}

\textbf{STGCNs}~\cite{yan2018spatial,zhou2023multi,zhou2021stgae} are widely adopted for skeleton-based human action analysis tasks, including action recognition and time-series forecasting. Typically, STGCNs model spatial features via graph convolution, while temporal dependencies are modeled in a separate stage. For spatial modeling, some methods~\cite{shi2019two,huang2020spatio} treat joints as graph nodes, while others~\cite{shi2019two,chen2021multi} additionally construct bone-based graphs. Since human actions often involve the coordinated motion of different limbs, part-level graph representations are effective for extracting semantic features~\cite{huang2020part}. For temporal modeling, temporal convolution and RNNs are commonly adopted. The key challenge is integrating spatial and temporal information effectively. Early approaches~\cite{yan2018spatial,shi2019skeleton,shi2019two} focused mainly on spatial modeling and incorporated temporal information separately. Some recent methods~\cite{liu2020disentangling,gao2023learning} construct 3D spatial-temporal graphs by combining consecutive spatial graphs, but this introduces substantial computational overhead. To address these issues, this work proposes a novel spatial-temporal module that injects temporal cues into graph convolution, enabling efficient spatial-temporal cohesion.

\textbf{Skeleton-Based Action Recognition}
aims to extract spatial–temporal patterns from skeletal data for action classification~\cite{ren2020survey}.
Early approaches mainly relied on handcrafted features, while recent deep learning methods have achieved superior performance, including both non-graph-based approaches~\cite{li2021memory,zhu2023motionbert,duan2022revisiting} and graph-based methods~\cite{li2019actional,huang2020part}.
Early non-graph-based methods \cite{duan2022revisiting} commonly represent skeleton sequences as ordered vectors or pseudo-images, and use RNNs, CNNs, or their variants to capture temporal and spatial patterns.
Meanwhile, hybrid designs have also been explored, where graph-based modules are used to model spatial joint relationships, while RNNs, temporal convolutions, or attention mechanisms are adopted to capture temporal dynamics. STGCN-based methods~\cite{shi2020skeleton,song2020stronger} further integrate graph convolution with temporal modeling modules, enabling explicit topology-aware spatial modeling together with temporal dependency learning.
Recent STGCN-based approaches further enhance spatial–temporal modeling through improved attention mechanisms \cite{yang2025restore_rwkv} or topology-aware designs~\cite{wu2025local,ma2025thtformer}.
Existing STGCN-based methods mainly improve performance by either deepening network architectures~\cite{yan2018spatial,shi2019skeleton} or widening them via multi-stream or disentangled designs~\cite{huang2020spatio,liu2020disentangling}.
However, deeper models often suffer from over-smoothing and increased training complexity, while wider architectures tend to emphasize coarse postural information rather than fine-grained motion patterns.
In this context, we propose an efficient dual-pathway architecture with dynamic fusion, which enables complementary modeling of spatial-temporal information.

\textbf{Dual-Pathway Networks} are inspired by the structure and function of the human visual system, which comprises the dorsal stream, responsible for spatial processing, and the ventral stream, involved in object recognition and perception. In dual-pathway networks, input data is processed independently by two parallel pathways before the extracted features are merged for downstream tasks. This architecture has been successfully applied in various domains, including computer vision \cite{chen2022dped,simonyan2014two,,yang2026unipet} and medical care \cite{shi2022attention,yang2024amir,wu2024attriprompter}.
In the realm of skeleton-based action recognition, dual-pathway networks \cite{shi2019two} or multi-pathway networks \cite{chen2021channel,liu2020disentangling} offer a promising approach for effectively integrating spatial and temporal cues, thereby enhancing performance and robustness. Recent evidence suggests that the dorsal and ventral pathways interact significantly, playing crucial roles in various aspects of cognition \cite{van2015interactions}. However, existing networks do not fully model these pathway interactions, limiting overall ensemble performance. To address this gap, we propose a novel dual-pathway network with dynamic fusion to fully leverage the complementary nature of skeletal data.

\MSsection{sec_empirical}{Problem Formulation and Empirical Analysis}

\myPara{Notations and Task Definition}
Skeleton-based action recognition aims to predict an action label from a sequence of human skeletons.
We denote a skeleton sequence as $\mathbf{X} \in \mathbb{R}^{T \times N \times C}$, where $T$ is the number of frames, $N$ is the number of joints, and $C$ is the feature dimension.
The skeletal structure is modeled as an undirected graph $\mathcal{G}=(\mathcal{V},\mathcal{E})$,
where nodes represent joints and each edge (bone) connects two joints.
The input features $\mathbf{X}$ are defined on the graph $\mathcal{G}$ and evolve over time.

\begin{figure}
    \centering
    \subfigure[Skeletal modality representations]{
        \includegraphics[width=0.49\linewidth,clip,trim=485 205 200 205]{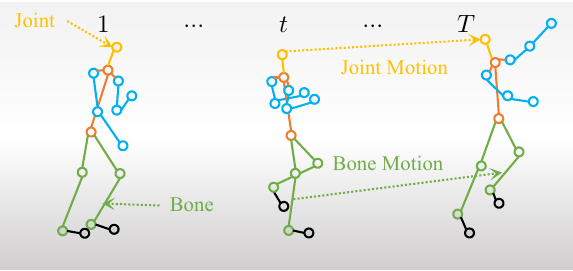}
        \label{fig:modality}
    }\subfigure[Dual-pathway inspiration]{
        \includegraphics[width=0.49\linewidth,clip,trim=200 205 485 205]{figs/brain-pathway.pdf}
        \label{fig:brain-pathway}
    }
    \subfigure[Modality imbalance and complementarity]{
        \includegraphics[width=0.49\linewidth]{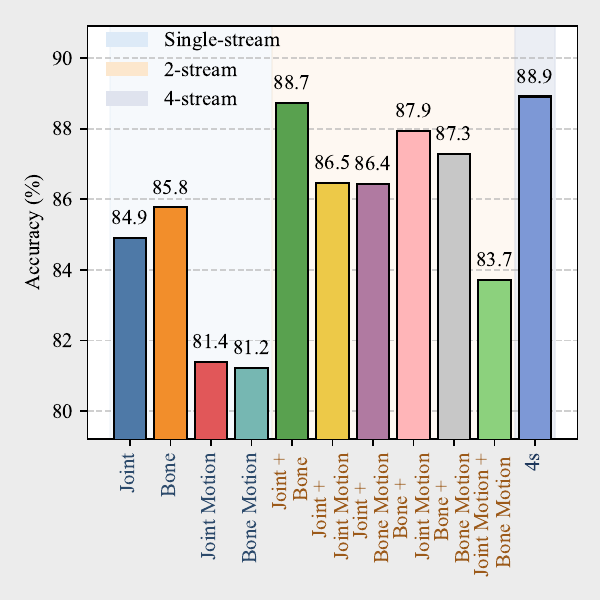}
        \label{fig:empirical}
    }\subfigure[Improved modality balance]{
        \includegraphics[width=0.49\linewidth]{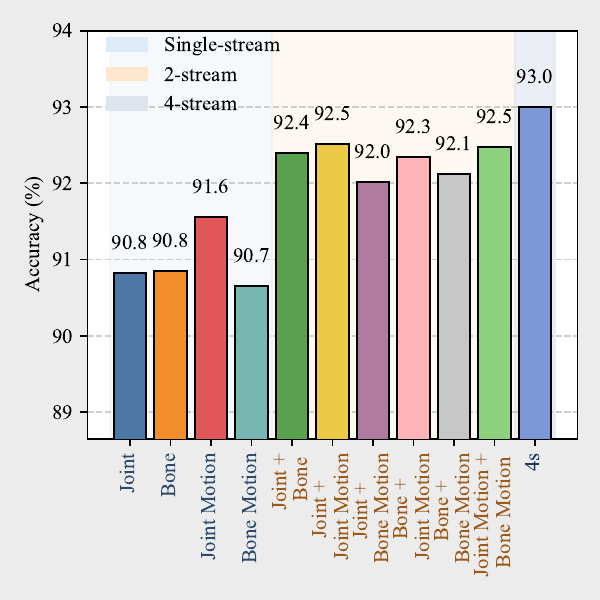}
        \label{fig:empirical-ours}
    }
    \MSfigcaption{fig:motivation}{
        \textbf{Motivation and empirical observations.}
        (a) Common skeletal modality representations derived from joint sequences, including joint positions, bone vectors, joint motion, and bone motion.
        (b) Inspiration from the two-stream hypothesis of the human visual system, where spatial structure and motion are processed by two parallel yet interacting pathways.
        (c) Empirical results of CTR-GCN~\cite{chen2021channel} on the X-Sub protocol of NTU RGB+D 120 under different skeletal modalities. We report Top-1 Accuracy (denoted as Accuracy). Single-stream results exhibit clear performance imbalance across modalities, while combining structural and motion information consistently improves performance, highlighting their complementarity.
        (d) Results of our method, which significantly alleviates modality imbalance and achieves more balanced performance across different skeletal modalities.
    }{}
\end{figure}

\myPara{Definition of the Skeletal Modality}
We define the \emph{skeletal modality} as a family of representations derived from skeletal data, which is distinct from other input modalities such as RGB or depth and is referred to as the \emph{modality} for brevity.
As illustrated in \cref{fig:modality}, skeletal modalities can be broadly categorized into \emph{spatial-aware modalities}, which encode human limb configurations using joint positions or bone vectors, and \emph{temporal-aware modalities}, which characterize motion dynamics through temporal variations of joint or bone representations.
In practice, we consider four commonly used skeletal modalities: \emph{joint position}, \emph{bone position}, \emph{joint motion}, and \emph{bone motion}.
Given the joint position sequence $\mathbf{X}_{\mathrm{jp}}$, the remaining modalities are deterministically derived via simple spatial and temporal differencing operations: joint motion is defined as the temporal difference of joint positions, bone position is obtained from the two joints connected by each bone, and bone motion is computed as the temporal difference of bone positions, which can be represented as:
\begin{equation}
    \begin{aligned}
        \mathbf{X}_{\mathrm{jm}}(t)   & = \mathbf{X}_{\mathrm{jp}}(t) - \mathbf{X}_{\mathrm{jp}}(t-1), \quad t = 2,\ldots,T, \\
        \mathbf{X}_{\mathrm{bp}}(t,e) & = \mathbf{X}_{\mathrm{jp}}(t, v_1(e)) - \mathbf{X}_{\mathrm{jp}}(t, v_2(e)),         \\
        \mathbf{X}_{\mathrm{bm}}(t)   & = \mathbf{X}_{\mathrm{bp}}(t) - \mathbf{X}_{\mathrm{bp}}(t-1), \quad t = 2,\ldots,T,
    \end{aligned}
\end{equation}
where $v_1(e)$ and $v_2(e)$ denote the two joints incident to bone $e$.
\myPara{Architectural Challenges}
From an architectural perspective, existing STGCNs can be broadly categorized into (2+1)D and 3D paradigms, depending on how spatial and temporal modeling are coupled within a single branch~\cite{yan2018spatial,shi2019two,liu2020disentangling}.
(2+1)D STGCNs~\cite{yan2018spatial,shi2019two} sequentially apply spatial graph convolution and temporal convolution, which is computationally efficient but enforces a rigid separation between spatial structure and temporal dynamics, limiting their interaction within a unified feature space.
In contrast, 3D STGCNs~\cite{gao2019optimized,liu2020disentangling} jointly model spatial and temporal dependencies by constructing spatiotemporal graphs over multiple frames, but the rapidly growing spatiotemporal neighborhood incurs high computational cost, thereby constraining the temporal receptive field and hindering long-range temporal modeling.
As a result, despite their distinct coupling strategies, both paradigms face inherent difficulties in achieving efficient spatial-temporal modeling for complex human actions.

\myPara{Empirical Study and Key Observations}
To examine how existing STGCNs exploit different skeletal modalities, we conduct an empirical analysis using CTR-GCN~\cite{chen2021channel} on the NTU RGB+D 120 X-Sub benchmark.
Following common practice~\cite{simonyan2014two}, different skeletal modalities are processed by independent streams, and ensemble performance is obtained by averaging prediction scores across multiple streams.
As shown in \cref{fig:empirical}, single-stream results exhibit a \textbf{clear performance imbalance}: joint and bone modalities achieve higher accuracy (84.91\% and 85.77\%) than motion-based modalities (81.40\% and 81.21\%).
This indicates that motion-only representations, which primarily encode temporal differences, lack sufficient spatial structure to reliably distinguish fine-grained actions when learned in isolation.
Two-stream fusion further reveals \textbf{modality complementarity}.
While combining structural modalities (joint + bone) yields a substantial improvement (88.73\%), fusing motion-only modalities (joint motion + bone motion) results in the lowest performance (83.71\%), suggesting that temporal cues alone are insufficient even when aggregated.
In contrast, integrating structural and motion modalities consistently improves accuracy, implying that motion information is most effective when grounded in explicit spatial structure.
These observations highlight that the imbalance across modalities is intrinsic rather than incidental, and that simply averaging predictions cannot fully resolve it.
Instead, they suggest that explicitly modeling the complementary spatial and temporal information within a unified representation is key to improving single-stream efficiency and reducing reliance on heavy multi-stream ensembles.

\myPara{Motivation of the Dual-Pathway Network}
Motivated by the above empirical findings, we seek a unified modeling framework that can explicitly exploit the complementary nature of spatial structure and temporal dynamics within a single network.
Our design is inspired by the two-stream hypothesis~\cite{goodale1992separate}, which posits that the human visual cortex processes spatial information through the ventral pathway and motion information through the dorsal pathway.
As illustrated in \cref{fig:brain-pathway}, these two pathways operate in parallel and interact to support coherent action perception.
Analogously, we introduce a dual-pathway architecture that separates spatial-aware and temporal-aware processing while enabling their collaboration for holistic action understanding.

\section{Dual-Pathway Graph Convolutional Networks (NeuroPath)}
In this section, we provide an overview of the proposed NeuroPath framework and subsequently present its core architectural components.

\subsection{Framework Overview}
\label{sec_motivation}

\begin{figure*}
    \centering
    \includegraphics[width=\linewidth]{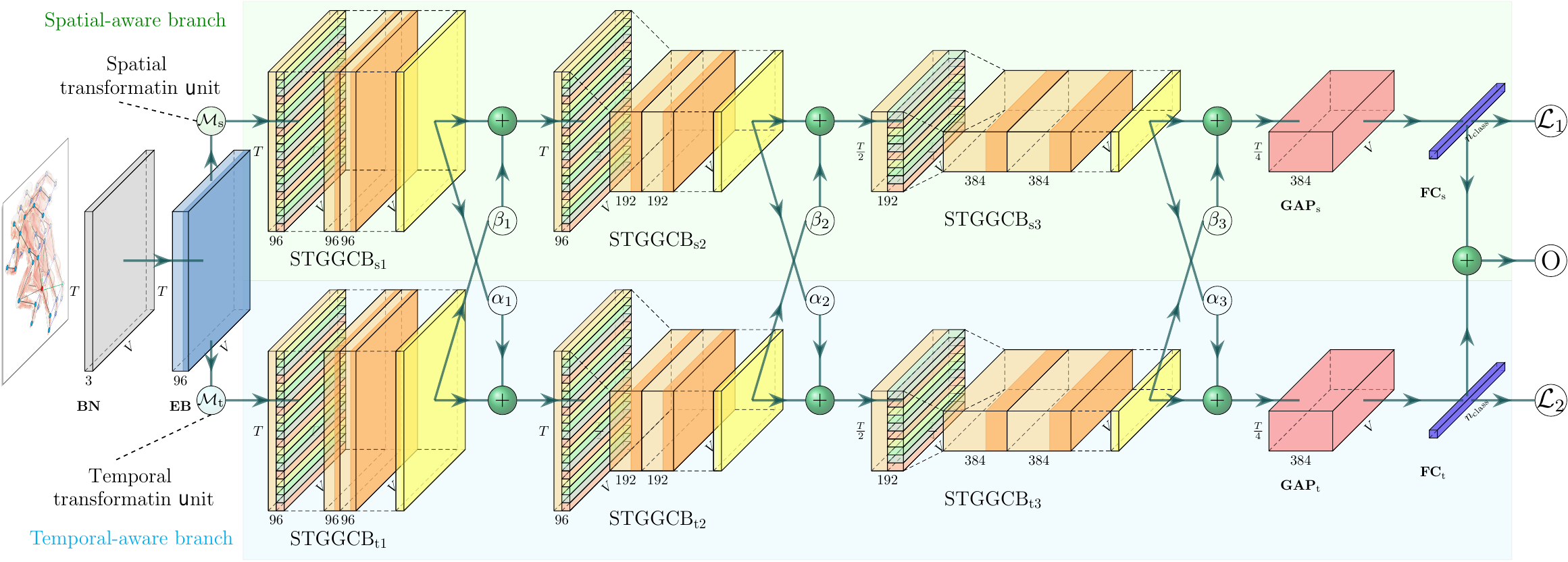}
    \caption{
        \textbf{Overview of the proposed NeuroPath framework.}
        NeuroPath adopts a brain-inspired dual-pathway architecture to explicitly model the complementary spatial-aware and temporal-aware information in skeletal data.
        The two pathways are conceptually motivated by the ventral and dorsal streams in the human visual system, which specialize in spatial structure and motion processing, respectively.
        In contrast to existing multi-stream networks that fuse modalities only at the output level, NeuroPath introduces dynamic inter-pathway fusion modules to enable effective interaction and coordination between spatial-aware and temporal-aware representations throughout the network.
    }
    \label{fig-framework}
\end{figure*}

\cref{fig-framework} illustrates the framework of NeuroPath, which is designed to explicitly exploit the complementary nature of spatial-aware and temporal-aware information.
Given the input skeletal sequence $\mathbf{X}$ defined on the graph $\mathcal{G}$,
NeuroPath first applies batch normalization (BN) and embedding (EB) to obtain a normalized feature representation $\tilde{\mathbf{X}} \in \mathbb{R}^{T \times N \times D}$, where $D$ denotes the embedding dimension.
Here, the embedding is implemented as a linear projection that maps the raw skeletal inputs into a unified feature space.
The normalized features $\tilde{\mathbf{X}}$ are then processed by two parallel pathways that focus on spatial-aware and temporal-aware representations, respectively.
Inspired by the functional segregation observed in the human visual system, the transformation units (see \cref{sec_trasformation}) decompose the input features into distinct skeletal modalities tailored to each pathway.
Subsequently, the Spatial--Temporal Group Graph Convolution Block (STGGCB, see \cref{sec_stggcb}) captures coordinated body-part movements by integrating spatial structure with temporal cues and modeling their interdependencies.
To facilitate effective information exchange between the two pathways, a dynamic fusion module (see \cref{sec_fusion}) is introduced to explicitly model inter-pathway interactions and enhance spatial--temporal feature integration.
Finally, the output features from both pathways are aggregated via a Global Average Pooling (GAP) layer and fed into a Fully Connected (FC) layer for action classification.
The entire network is supervised using pathway-specific cross-entropy losses, denoted as $\mathcal{L}_1$ and $\mathcal{L}_2$, which are applied to the spatial-aware and temporal-aware pathways, respectively, enabling stable and collaborative learning of spatial–temporal representations.

\subsection{Spatial/Temporal Transformation Unit}
\label{sec_trasformation}

\myPara{Design Rationale}
As discussed in \cref{sec_empirical}, existing STGCN-based approaches commonly exploit multiple skeletal modalities through independent streams and perform fusion at the score level~\cite{shi2019skeleton,liu2020disentangling,Wang_2023_CVPR,chen2021channel}.
While this strategy empirically demonstrates modality complementarity, it also exposes two inherent limitations.
First, different modalities exhibit pronounced performance imbalance, causing motion-dominated streams to contribute marginally in isolation.
Second, treating modalities independently prevents explicit coordination between spatial structure and temporal dynamics during representation learning, relegating their interaction to late-stage fusion.
Our objective is fundamentally different.
Rather than introducing additional streams or relying on ensemble strategies, we aim to explicitly coordinate complementary spatial and temporal information \emph{within a single unified network}.
To this end, we introduce a \emph{spatial/temporal transformation unit}, which decomposes the input skeletal representation into spatial-aware and temporal-aware components and routes them to two dedicated pathways.
This design directly addresses the observed imbalance by ensuring that each pathway receives modality-consistent information, while enabling subsequent layers to model their interaction explicitly.

\myPara{Modality Transformation}
Following the definition in \cref{sec_empirical}, we consider four skeletal modalities,
denoted by
$\mathcal{X} = \{\mathbf{X}_{\mathrm{jp}}, \mathbf{X}_{\mathrm{bp}},
    \mathbf{X}_{\mathrm{jm}}, \mathbf{X}_{\mathrm{bm}}\}$,
corresponding to joint position, bone position, joint motion, and bone motion, respectively.
All four modalities are computed beforehand from the joint-position sequence
and are simultaneously available during training and inference.
Joint and bone positions encode spatial structure,
while joint and bone motion characterize temporal dynamics.
Let $\tilde{\mathbf{X}}$ denote the normalized skeletal features and
$\phi \in \mathcal{X}$ denote the \emph{modality identifier} of the current stream setting.
Importantly, $\phi$ does \emph{not} indicate that one modality is derived from another.
Instead, it specifies how the pre-computed modalities are \emph{paired and routed}
to the spatial-aware and temporal-aware pathways.
We define two deterministic, parameter-free transformation functions,
$\mathcal{M}_{\mathrm{s}}(\cdot)$ and $\mathcal{M}_{\mathrm{t}}(\cdot)$,
which select the spatial-aware and temporal-aware representations, respectively.
These functions do not perform any numerical transformation;
they only determine which modality is assigned to each pathway.
Accordingly, the spatial-aware representation is defined as
\begin{equation} \label{eq_ms}
    \mathbf{X}_{\mathrm{s}} = \mathcal{M}_{\mathrm{s}}(\phi) =
    \mathbb{I}_{\mathrm{jp}}(\phi)\,\mathbf{X}_{\mathrm{jp}} +
    \mathbb{I}_{\mathrm{bp}}(\phi)\,\mathbf{X}_{\mathrm{bp}},
\end{equation}
and the temporal-aware representation is defined as
\begin{equation} \label{eq_mt}
    \mathbf{X}_{\mathrm{t}} =  \mathcal{M}_{\mathrm{t}}(\phi) =
    \mathbb{I}_{\mathrm{jm}}(\phi)\,\mathbf{X}_{\mathrm{jm}} +
    \mathbb{I}_{\mathrm{bm}}(\phi)\,\mathbf{X}_{\mathrm{bm}},
\end{equation}
where $\mathbb{I}_{(\cdot)}(\phi)$ is an indicator function that equals $1$
if modality $\phi$ corresponds to the specified representation and $0$ otherwise.

\myPara{Discussion and Benefits}
This routing strategy does not introduce additional parameters and does not increase model expressivity by itself.
Instead, it provides a structured decomposition that exposes complementary spatial and temporal cues to different pathways, facilitating more effective coordination in subsequent modules.
By explicitly separating spatial structure and temporal dynamics at the input stage, the proposed transformation unit alleviates the imbalance observed in conventional multi-stream settings and lays the foundation for coherent spatial-temporal modeling.

\subsection{Spatial-Temporal Group Graph Convolution Block}
\label{sec_stggcb}

\begin{figure}
    \centering
    \includegraphics[width=\linewidth]{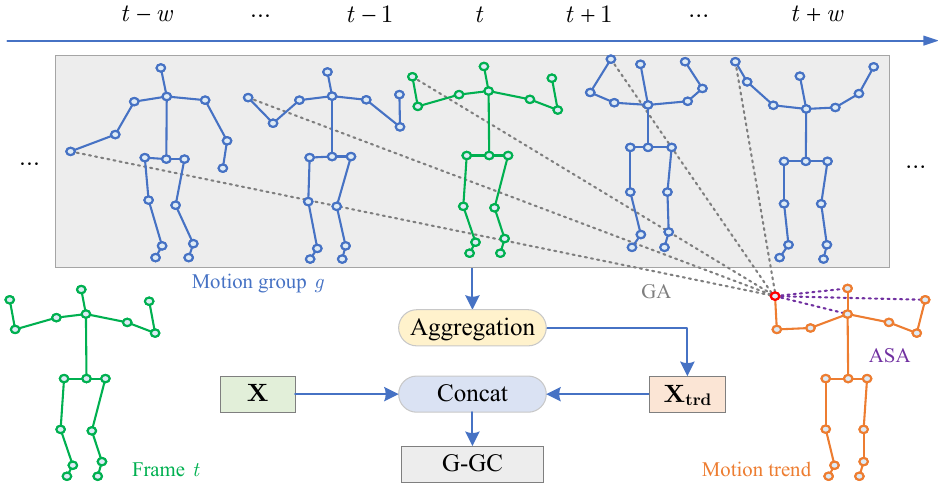}
    \caption{\textbf{Core design intuition of STGGCB}.
        Discriminative actions are characterized by coordinated groups of joints with correlated temporal dynamics.
        STGGCB first summarizes local temporal variation at the group level and then conditions spatial aggregation on this motion-aware context, enabling graph convolution to emphasize structurally and temporally relevant joint interactions.}
    \label{fig-main_idea}
\end{figure}

\begin{figure}
    \centering
    \subfigure{\label{fig-ST_conv-a}}
    \subfigure{\label{fig-ST_conv-b}}

    \includegraphics[width=\linewidth]{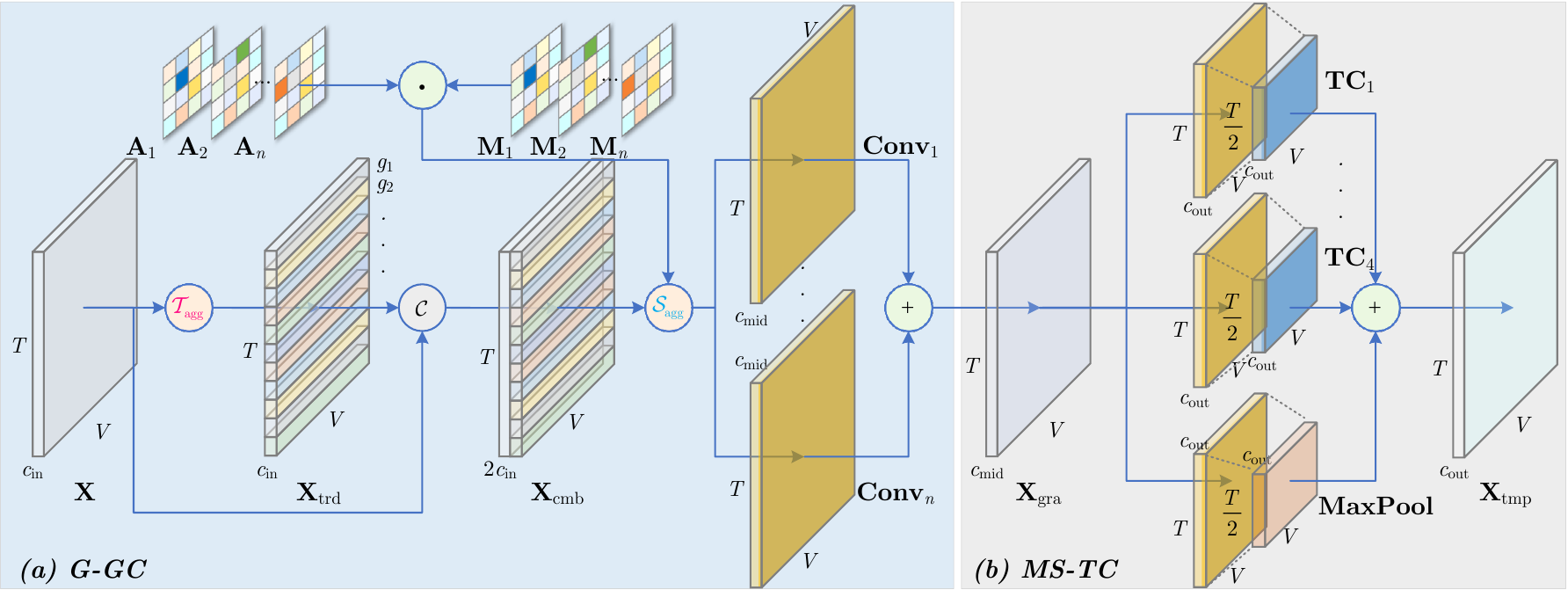}
    \caption{
        \textbf{The network architecture of STGGCB}: (a) Group Graph Convolution (G-GC) and (b) Multi-Scale Temporal Convolution (MS-TC). G-GC comprises group aggregation (see \cref{eq_tagg}) and adaptive structural aggregation (see \cref{eq_gra}).
    }
    \label{fig-ST_conv}
\end{figure}

\myPara{Design Rationale}
As illustrated in \cref{fig-main_idea}, a key challenge in skeleton-based action recognition
is modeling coordinated body-part interactions that evolve over time.
Discriminative actions are often characterized not by isolated joint motions,
but by structured groups of joints exhibiting correlated temporal dynamics
(e.g., arm–torso coordination in waving or punching).
However, most existing (2+1)D STGCNs~\cite{yan2018spatial,shi2019skeleton,shi2019two}
either decouple spatial and temporal modeling
or rely on sequential processing,
which limits the ability of spatial aggregation to condition on temporal variation.
To address this limitation, we design the Spatial-Temporal Group Graph Convolution Block (STGGCB, see \cref{fig-ST_conv})
with two core objectives:
(1) to explicitly inject temporal variation cues into spatial aggregation,
so that graph convolution can prioritize motion-relevant joint interactions; and
(2) to capture multi-hop structural dependencies while avoiding redundant message mixing
that commonly arises in dense spatial-temporal graphs.
STGGCB integrates temporal-aware group aggregation with adaptive structural aggregation,
followed by efficient temporal context modeling and spatial–temporal attention,
thereby enabling coherent and scalable spatial–temporal feature learning within each pathway.

\myPara{Group Aggregation}
As shown in \cref{fig-ST_conv-a}, group aggregation is designed to explicitly inject temporal evolution cues into spatial graph convolution, enabling motion-aware neighborhood aggregation.
Instead of relying on a separate temporal convolution to capture motion patterns,
we first compute a local temporal evolution feature that characterizes short-term motion trends of joints,
and concatenate it with the original feature to form a temporally enhanced representation:
\begin{equation} \label{eq_tagg}
    \mathbf{X}_{\mathrm{cmb}}^{(l)} =
    \mathrm{Concat}\!\left(\mathbf{X}^{(l)}, \mathbf{X}^{(l)}_{\mathrm{trd}}\right), \quad
    \mathbf{X}^{(l)}_{\mathrm{trd}} = \mathcal{T}_{\mathrm{agg}}\!\left(\mathbf{X}^{(l)}\right),
\end{equation}
where $\mathcal{T}_{\mathrm{agg}}$ is implemented as a second-order temporal difference operator,
which emphasizes motion transitions while suppressing constant-velocity patterns.
By embedding temporal variation directly into the features used for graph convolution,
group aggregation allows spatial message passing to prioritize motion-relevant neighbors,
thereby achieving tighter spatial--temporal coupling than conventional $(2{+}1)$D designs
that model spatial and temporal information sequentially.

\begin{figure}
    \centering
    \includegraphics[height=\linewidth]{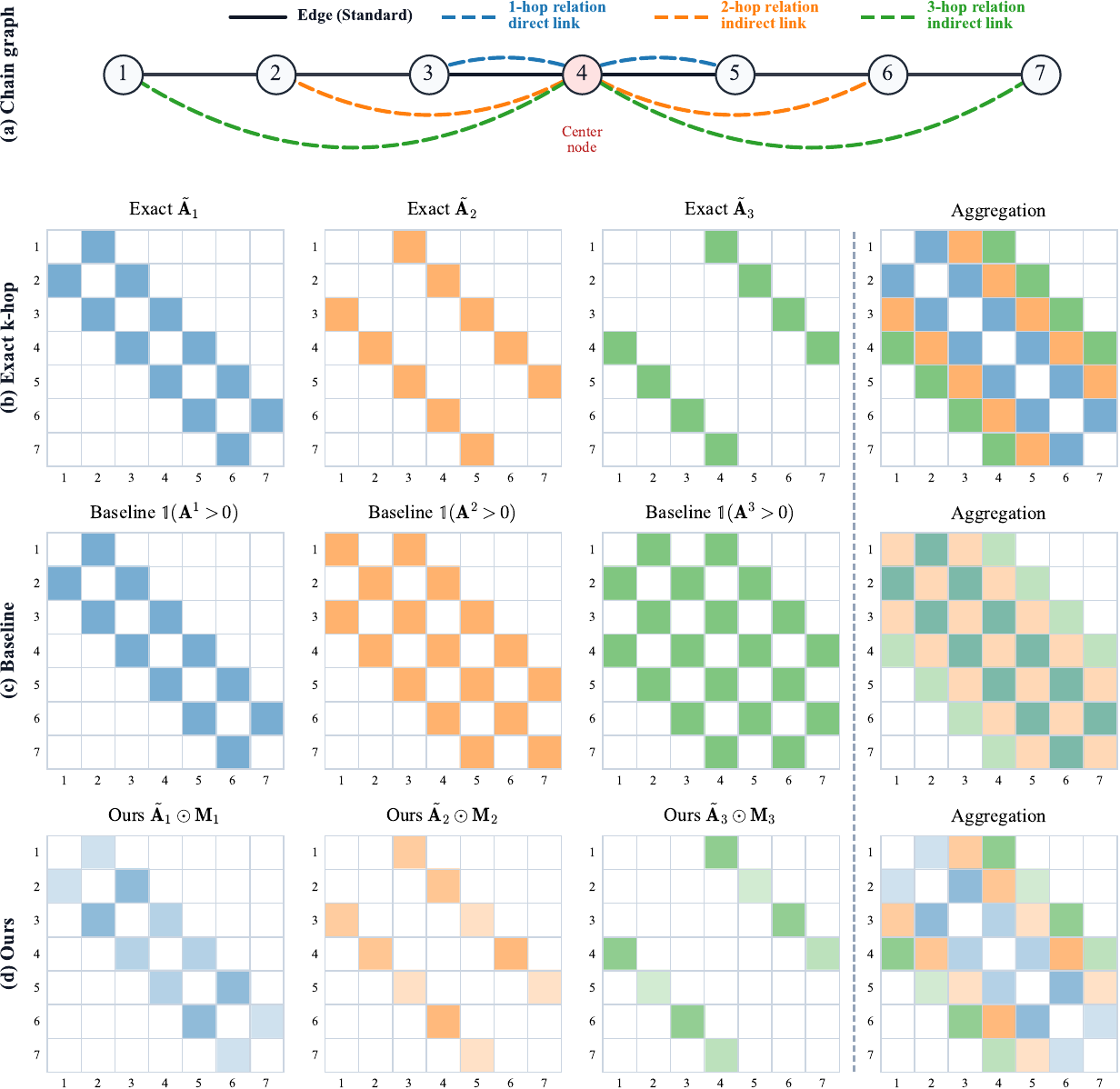}
    \subfigure{\label{fig-decomposition-a}}
    \subfigure{\label{fig-decomposition-b}}
    \subfigure{\label{fig-decomposition-c}}
    \subfigure{\label{fig-decomposition-d}}
    \MSfigcaption{fig-decomposition}{
        \textbf{Illustration of multi-hop aggregation strategies on a simple chain graph.}
        (a) A 7-node chain graph, where node 4 is selected as the center node and 1-hop, 2-hop, and 3-hop relations are shown by different colors.
        (b) Exact $k$-hop adjacency supports $\tilde{A}_k$, where each matrix contains only node pairs whose shortest-path distance is exactly $k$.
        (c) Conventional baselines using cumulative multi-hop aggregation ($A^1 + A^2 + A^3$), which entangle different hop distances within a single representation.
        (d) Our method, which applies learnable masks $M_k$ over exact $k$-hop supports, enabling adaptive reweighting while preserving explicit separation of neighborhood ranges.
    }{
    }
\end{figure}

\myPara{Adaptive Structural Aggregation}
To capture long-range spatial dependencies in a controlled manner, we introduce an adaptive structural aggregation that reweights multi-hop message passing without modifying graph connectivity, as shown in \cref{fig-ST_conv-a}.
Given the combined features $\mathbf{X}_{\mathrm{cmb}}^{(l)}$, aggregation over $K$-hop neighborhoods is formulated as
\begin{align}
    \mathbf{X}^{(l)}_{\mathrm{gra}} & = \sigma\!\left(\sum_{k=0}^{K}\bar{\mathbf{A}}_{k}\,\mathbf{X}^{(l)}_{\mathrm{cmb}}\,\mathbf{W}^{(l)}_{k}\right), \label{eq_gra} \\
    \bar{\mathbf{A}}_{k}            & = \tilde{\mathbf{D}}^{-\frac{1}{2}}\!\left(\tilde{\mathbf{A}}_{k}\odot \mathbf{M}_{k}\right)\tilde{\mathbf{D}}^{-\frac{1}{2}},
\end{align}
where $\tilde{\mathbf{A}}_{k}$ denotes the fixed \emph{exact $k$-hop adjacency support},
i.e., $\tilde{\mathbf{A}}_{k}(i,j)=1$ if and only if nodes $i$ and $j$ are connected by
a path of length exactly $k$ in the original skeleton graph, and $0$ otherwise.
This definition ensures that only relations at hop $k$ are aggregated at the $k$-th term,
avoiding redundant mixing across different neighborhood ranges.
The matrix $\mathbf{M}_{k}$ is a learnable strength mask that adaptively reweights
existing $k$-hop connections.
The element-wise modulation $\odot$ adjusts the influence of existing multi-hop connections while preserving the original topology, thereby avoiding redundant shortcut edges that commonly arise from additive adjacency updates on dense $k$-hop graphs.
The mask $\mathbf{M}_{k}$ is generated by a cross-attention gate,
\begin{equation} \label{eq_mk}
    \mathbf{M}_{k}=\mathbf{B}_{k}+\mathrm{Softmax}\!\left(\frac{\mathbf{Q}\mathbf{K}^{\top}}{\sqrt{d}}\right),
\end{equation}
where $\mathbf{Q}$ and $\mathbf{K}$ are feature embeddings from $\mathbf{X}^{(l)}_{\mathrm{cmb}}$ and $\mathbf{X}^{(l)}$, respectively, and $\mathbf{B}_{k}$ provides a hop-specific prior that stabilizes structural weighting when the normalized attention becomes diffuse.
Unlike conventional higher-order topological message passing (see \cref{fig-decomposition-c}) that explicitly alters graph connectivity, this formulation preserves the original multi-hop supports and enables adaptive long-range aggregation with controlled complexity and without increasing graph density (see \cref{fig-decomposition-d}).
\myPara{Multi-Scale Temporal Convolution}
As shown in \cref{fig-ST_conv-b}, we adopt a standard multi-scale temporal convolution to complement spatial–temporal aggregation.
Specifically, the input channels are evenly divided into $g$ groups, and each group is processed independently.
The resulting features are concatenated as
\begin{equation}
    \mathbf{X}^{(l)}_{\mathrm{tmp}} =
    \mathrm{Concat}\!\left(
    \mathrm{Conv}_{d}\!\left(\mathbf{X}^{(l)}_{\mathrm{gra}}\right),
    \mathrm{MP}\!\left(\mathbf{X}^{(l)}_{\mathrm{gra}}\right)
    \right),
\end{equation}
where $\mathrm{Conv}_{d}(\cdot)$ denotes a dilated temporal convolution and $\mathrm{MP}(\cdot)$ denotes temporal max pooling.
This module enlarges the temporal receptive field with low computational overhead and serves as a conventional temporal modeling component.

\begin{figure}
    \centering
    \includegraphics[width=\linewidth]{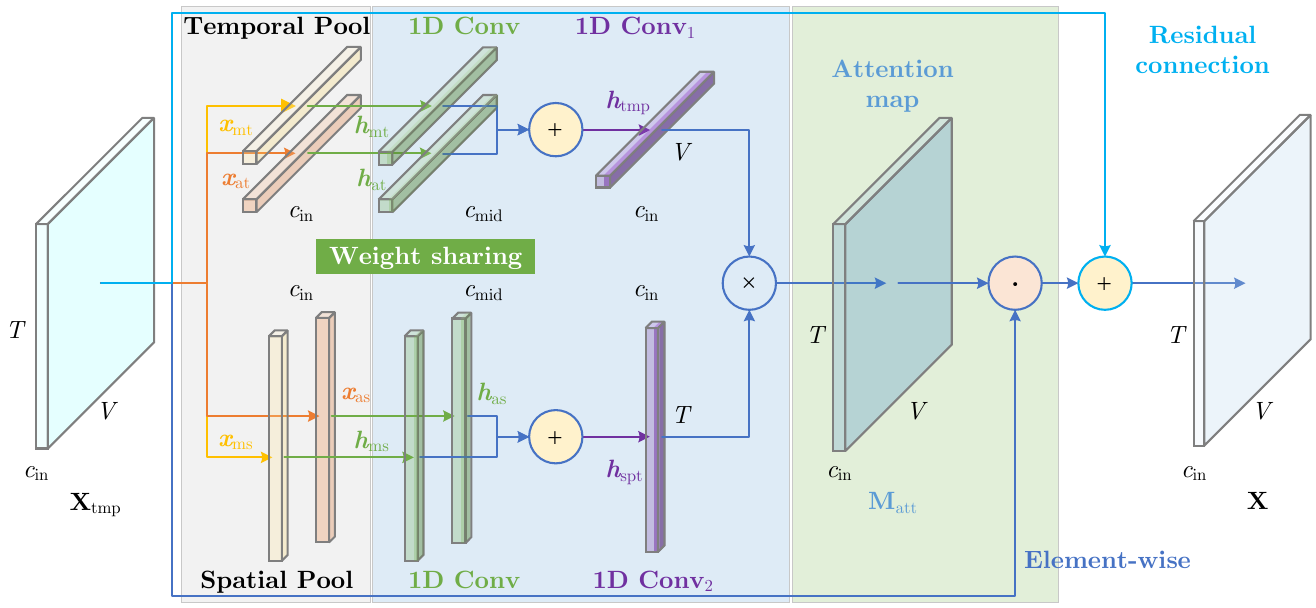}
    \caption{
        \textbf{The illustration of the spatial-temporal attention module.} `$\cdot$' and `$\times$' represent matrix dot-product and matrix multiplication operations, respectively.
    }
    \label{fig-ST_att}
\end{figure}

\myPara{Spatial-Temporal Attention} \label{sec_st_att}
To alleviate potential over-smoothing caused by repeated spatial–temporal aggregation,
we incorporate a spatial–temporal attention module to adaptively reweight features.
Given the input feature $\mathbf{X}_{\mathrm{in}}$, we summarize global context along the spatial
and temporal dimensions using both average pooling and max pooling.
Specifically, we obtain pooled features
$\bm{x}_{\mathrm{ms}}, \bm{x}_{\mathrm{as}} \in \mathbb{R}^{C_{\mathrm{in}} \times T}$ and
$\bm{x}_{\mathrm{mt}}, \bm{x}_{\mathrm{at}} \in \mathbb{R}^{C_{\mathrm{in}} \times N}$,
which capture complementary statistics of the feature distribution.
These pooled features are projected by shared 1D convolutions to produce
\begin{equation}
    \bm{h}_{\mathrm{ms}},\bm{h}_{\mathrm{as}},\bm{h}_{\mathrm{mt}},\bm{h}_{\mathrm{at}}
    = \sigma\!\left(\mathrm{Conv}_{1D}\!\left(
        \bm{x}_{\mathrm{ms}},\bm{x}_{\mathrm{as}},\bm{x}_{\mathrm{mt}},\bm{x}_{\mathrm{at}}
        \right)\right).
\end{equation}
The decoded spatial and temporal attention signals are then combined to form
a spatial–temporal attention map
\begin{equation} \label{eq_att}
    \mathbf{M}_{\mathrm{att}} =
    \kappa\!\left(\mathrm{Conv}_{1D}\!\left(\bm{h}_{\mathrm{ms}}+\bm{h}_{\mathrm{as}}\right)\right)
    \times
    \kappa\!\left(\mathrm{Conv}_{1D}\!\left(\bm{h}_{\mathrm{mt}}+\bm{h}_{\mathrm{at}}\right)\right),
\end{equation}
where $\kappa(\cdot)$ denotes the sigmoid function.
Finally, the attention map is applied to refine the features via a residual formulation:
\begin{equation}
    \mathbf{X}^{(l+1)} =
    \sigma\!\left(\mathrm{BN}\!\left(
        \mathbf{X}_{\mathrm{tmp}} \odot \mathbf{M}_{\mathrm{att}} + \mathbf{X}_{\mathrm{tmp}}
        \right)\right).
\end{equation}
This module follows common attention designs and serves as a lightweight feature reweighting mechanism
to stabilize training and preserve discriminative information.

\myPara{Discussion and Benefits}
STGGCB enables effective long-range spatial-temporal modeling without modifying graph topology or increasing message-passing order.
By injecting temporal variation into spatial aggregation and adaptively reweighting fixed multi-hop supports,
the block captures global joint coordination while avoiding redundant connections and oversmoothing.
Rather than increasing expressivity in theory, STGGCB provides a structured inductive bias
that exposes complementary spatial and temporal cues, leading to more stable and efficient learning in practice.

\begin{figure}
    \centering
    \subfigure{\label{fig-architecture-a}}
    \subfigure{\label{fig-architecture-b}}
    \subfigure{\label{fig-architecture-c}}
    \includegraphics[width=\linewidth]{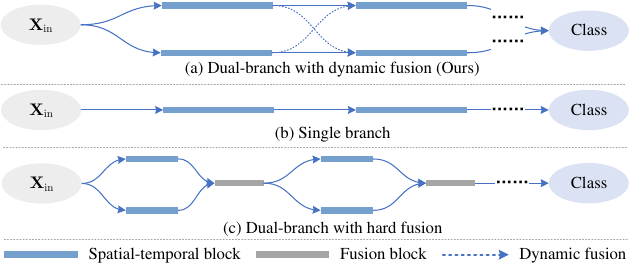}
    \caption{
        A comparison of STGCN network architectures: (a) our dual-pathway with dynamic fusion, (b) single pathway, (c) dual-pathway with hard fusion.
    }
    \label{fig-architecture}
\end{figure}

\subsection{Dual-Pathway Fusion}
\label{sec_fusion}

\myPara{Design Rationale}
\cref{fig-architecture} contrasts three paradigms.
The single-branch STGCN (see \cref{fig-architecture-b}) \cite{yan2018spatial,shi2019two,chen2021channel} implicitly entangles spatial structure and temporal dynamics in a shared representation, which limits their distinct contributions.
A dual-branch design with hard fusion (see \cref{fig-architecture-c}) \cite{liu2020disentangling,huang2020spatio,zhang2020sta} separates spatial and temporal processing but combines them using fixed operators, preventing adaptive interaction during representation learning.
In contrast, our dual-branch architecture with dynamic fusion (see \cref{fig-architecture-a} explicitly separates spatial-aware and temporal-aware representations while enabling controlled, stage-wise interaction throughout the network.
This mechanism fundamentally differs from conventional mid-level multimodal fusion, which typically merges heterogeneous features once at a predefined layer.
Instead, NeuroPath enforces progressive coordination across the representation hierarchy, allowing spatial and temporal cues to influence each other during feature formation.

\myPara{Dynamic Inter-Pathway Fusion}
We perform \emph{dynamic fusion} at multiple depths of the network.
Specifically, after each STGGCB, features from the two pathways are softly exchanged,
allowing complementary information to be progressively integrated while preserving pathway specialization.
Formally, let $\mathbf{X}_{\mathrm{s}}$ and $\mathbf{X}_{\mathrm{t}}$ denote the spatial- and temporal-pathway features at a given stage.
The fusion operation is defined as
\begin{equation}
    \mathbf{X}'_{\mathrm{s}} = \mathbf{X}_{\mathrm{s}} + \alpha\,\mathbf{X}_{\mathrm{t}}, \qquad
    \mathbf{X}'_{\mathrm{t}} = \mathbf{X}_{\mathrm{t}} + \beta\,\mathbf{X}_{\mathrm{s}},
\end{equation}
where $\alpha$ and $\beta$ are learnable scalar weights.
This formulation implements a lightweight residual exchange mechanism,
enabling each pathway to selectively absorb complementary cues from the other
without collapsing them into a single representation.
The fused features $\mathbf{X}'_{\mathrm{s}}$ and $\mathbf{X}'_{\mathrm{t}}$ are then propagated to the next block.

\myPara{Discussion and Benefits}
Compared with multi-stream ensemble methods~\cite{chen2021channel}, which fuse independently trained networks only at the score level, our fusion operates within a single network and during feature learning.
This enables spatial and temporal cues to interact continuously across the representation hierarchy, rather than compensating for each other only at the final prediction stage.
Importantly, the proposed fusion does not aim to increase expressivity through complex operators; instead, it introduces a structured coordination mechanism that explicitly exposes the imbalanced yet complementary nature of spatial and temporal information.
As a result, NeuroPath achieves more effective and stable spatial–temporal coordination, reducing reliance on heavy multi-stream ensembles while improving accuracy and robustness in practice.

\section{Experiments} \label{sec:exp}
This section presents the experimental setup followed by the experimental results.

\subsection{Experimental Setting}
\myPara{Datasets}
\textbf{Kinetics Skeleton 400} \cite{carreira2017quo} is obtained from YouTube by OpenPose \cite{cao2019openpose}, containing 260,232 sequences with over
400 action types. The training and testing sets have 240,436 and 19,796 sequences, respectively. Each skeleton has 18 joints with 2D coordinates and prediction confidence as initial features. Two skeletons with higher overall confidence scores are preserved.  Each sequence is padded to 300 by repeating it. For the Kinetics Skeleton 400 dataset, we report Top-1 and Top-5 accuracies for evaluation.
\textbf{NTU RGB+D 60} \cite{shahroudy2016ntu} is collected by Microsoft Kinetics v2, including 56,578 sequences with over 60 action classes from 40 subjects and 3 camera views.
Each skeleton contains 25 joints with 3D coordinates.
The number of subjects in each frame ranges from 1 to 2.
Two evaluation protocols are used: Cross-Subject (X-Sub) and Cross-View (X-View).
In X-Sub, 40,091 samples from 20 subjects are used for training and 16,487 for testing.
In X-View, 37,646 samples from camera 1 are used for training and 18,932 for testing.
\textbf{NTU RGB+D 120} \cite{liu2019ntu} extends the above with 57,367 samples with over 60 action classes, forming 113,945 samples of over 120 classes from 106 subjects and 32 camera setups.
It replaces X-View with Cross-Setup (X-Set), where 54,468 samples from 16 setups are used for training and 59,477 for testing.
In X-Sub, 63,026 samples from 53 subjects are for training and 50,919 for testing.

\myPara{Multi-Stream Evaluation Protocol}
Following common practice \cite{shi2019two,chen2021channel} in skeleton-based action recognition,
we also report multi-stream results by combining predictions
from multiple modality-specific inputs (e.g., joint, bone, joint motion, bone motion).
These multi-stream results are used \emph{only for evaluation purposes}
to facilitate fair comparison with prior ensemble-based methods.
Importantly, NeuroPath itself is trained end-to-end as a single unified model.
The reported 4-stream results are obtained by averaging the prediction scores
of independently evaluated NeuroPath models under different input modalities,
and do not affect the training procedure.

\myPara{Implementation Details}
We implement NeuroPath using PyTorch and train it on two GeForce RTX 3090 GPUs.
It is trained using SGD with a momentum of 0.9 and a batch size of 64.
The initial learning rate is set to 0.1 for 70 epochs with step learning rate decay with a factor of 0.1 at epochs \{30, 50\}. Weight decay is set to 0.0001 and 0.0005 for Kinetics Skeleton 400 and NTU RGB 60 \& 120. The processing of Kinetics Skeleton 400 is in line with \cite{liu2020disentangling}, and NTU-RGB+D 60 and 120 are based on \cite{chen2021channel}.
To speed up the training process on Kinetics Skeleton 400, 64 frames are randomly selected for the first 30 epochs and are successively increased to 300. For NTU RGB+D 60 and 120, 64 frames are selected, same as \cite{chen2021channel}.
\begin{table*}[]
    \centering
    \MStabcaption{tab-compare}{
        Accuracy (\%) comparison.
        The symbol ``--'' denotes unreported results.
        Underlined values indicate results taken from prior work~\cite{gao2023learning}.
        Bold values indicate the best result in each column.
    }{}
    \resizebox{\textwidth}{!}{
        \begin{tabular}{clrcccccccc}
            \toprule
            \multirow{2}{*}[-0.8ex]{ \bf \shortstack{ Ensemble                                                                                                                                                                                              \\ stream} } &
            \multirow{2}{*}[-0.8ex]{ \bf Method }           &
            \multirow{2}{*}[-0.8ex]{ \bf Publication }      &
            \multirow{2}{*}[-0.8ex]{ \bf \shortstack{Model                                                                                                                                                                                                  \\ type} } &
            \multicolumn{2}{c}{ \bf Kinetics Skeleton 400 } &
            \multicolumn{2}{c}{ \bf NTU RGB+D 60 }          &
            \multicolumn{2}{c}{ \bf NTU RGB+D 120 }                                                                                                                                                                                                         \\
            \cmidrule(lr){5-6} \cmidrule(lr){7-8} \cmidrule(lr){9-10}
                                                            &                                                &             &              & Top-1            & Top-5            & X-Sub            & X-View           & X-Sub            & X-Setup          \\
            \midrule
            \multirow{12}{*}[-1.8ex]{ \shortstack{Joint-                                                                                                                                                                                                    \\only} }
                                                            & ST-GCN \cite{yan2018spatial}                   & AAAI 2018   & (2+1)D       & 30.7             & 52.8             & 81.5             & 88.3             & 70.7             & 73.2             \\
                                                            & AGCN \cite{shi2019two}                         & CVPR 2019   & (2+1)D       & 34.8             & 56.5             & 86.8             & 94.2             & 77.9             & 78.5             \\
                                                            & Shift-GCN \cite{cheng2020skeleton}             & CVPR 2020   & (2+1)D       & \underline{34.7} & \underline{56.8} & 87.8             & 95.1             & 80.9             & 83.2             \\
                                                            & CTR-GCN \cite{chen2021channel}                 & ICCV 2021   & (2+1)D       & \underline{36.5} & \underline{58.5} & \underline{89.0} & \underline{94.8} & 83.6             & 84.5             \\
                                                            & InfoGCN \cite{chi2022infogcn}                  & CVPR 2022   & (2+1)D       & \underline{36.7} & \underline{58.7} & \underline{90.4} & \underline{95.0} & \underline{84.1} & \underline{85.2} \\
                                                            & ML-STGNet \cite{zhu2022multilevel}             & TIP 2022    & (2+1)D       & --               & --               & 89.8             & 95.0             & 84.9             & 86.5             \\
                                                            & MS-G3D \cite{liu2020disentangling}             & CVPR 2020   & (2+1)D \& 3D & \underline{36.5} & \underline{58.6} & 89.4             & 95.0             & 83.1             & 85.0             \\
                                                            & HetGCN \cite{gao2023learning}                  & TNNLS 2024  & 3D           & \textbf{37.1}    & 59.2             & 90.5             & 95.3             & 84.3             & 85.7             \\
                                                            & TranSkeleton \cite{liu2023transkeleton}        & TCSVT 2023  & Transformer  & --               & --               & 90.1             & 95.4             & 84.9             & 86.3             \\
                                                            & MotionBERT (scratch)~\cite{zhu2023motionbert}  & ICCV 2023   & Transformer  & --               & --               & 87.7             & 94.1             & --               & --               \\
                                                            & MotionBERT (finetune)~\cite{zhu2023motionbert} & ICCV 2023   & Transformer  & --               & --               & \textbf{93.0}    & \textbf{97.2}    & --               & --               \\
            \cline{2-10}
                                                            & NeuroPath (Ours)                               & --          & (2+1)D       & 37.0             & \textbf{59.8}    & {90.8}           & {95.8}           & \textbf{85.2}    & \textbf{87.4}    \\
            \midrule \multirow{18}{*}[-1.8ex]{ \shortstack{Multi-                                                                                                                                                                                           \\stream}}
                                                            & 4s AAGCN \cite{shi2020skeleton}                & TIP 2020    & (2+1)D       & 37.8             & 61.0             & 90.0             & 96.2             & --               & --               \\
                                                            & 4s Shift-GCN \cite{cheng2020skeleton}          & CVPR 2020   & (2+1)D       & 37.1             & 60.0             & 89.7             & 96.0             & 85.3             & 86.6             \\
                                                            & 3s Hyper-GNN \cite{hao2021hypergraph}          & TIP 2021    & (2+1)D       & 37.1             & 60.0             & 89.5             & 95.7             & --               & --               \\
                                                            & 4s MST-GCN \cite{chen2021multi}                & AAAI 2021   & (2+1)D       & 37.8             & 60.3             & 91.1             & 96.4             & 87.0             & 88.3             \\
                                                            & 4s CTR-GCN \cite{chen2021channel}              & ICCV 2021   & (2+1)D       & \underline{37.8} & \underline{60.2} & 92.4             & 96.0             & 88.7             & 90.1             \\
                                                            & 6s InfoGCN \cite{chi2022infogcn}               & CVPR 2022   & (2+1)D       & --               & --               & 93.0             & 97.1             & 89.8             & 91.2             \\
                                                            & 2s ML-STGNet \cite{zhu2022multilevel}          & TIP 2023    & (2+1)D       & 38.9             & 62.2             & 91.9             & 96.2             & 88.6             & 90.0             \\
                                                            & 4s KP-CTRGCN \cite{Wang_2023_CVPR}             & CVPR 2023   & (2+1)D       & --               & --               & 92.9             & 96.8             & 90.0             & 91.3             \\
                                                            & 4s FRGCN \cite{zhou2023learning}               & CVPR 2023   & (2+1)D       & --               & --               & 92.8             & 96.8             & 89.5             & 90.9             \\
                                                            & 4s JMDA \cite{xiang2025joint}                  & TOMM 2025   & (2+1)D       & --               & --               & 92.9             & 96.6             & 89.2             & 90.9             \\
                                                            & 4s LG-SGNet \cite{wu2025local}                 & PR 2025     & (2+1)D       & --               & --               & \textbf{93.1}    & 96.7             & 89.4             & 91.0             \\
                                                            & 2s MS-G3D \cite{liu2020disentangling}          & CVPR 2020   & (2+1)D \& 3D & 38.0             & 60.9             & 91.5             & 96.2             & 86.9             & 88.4             \\
                                                            & 4s DualHead \cite{chen2021learning}            & ACM MM 2021 & (2+1)D \& 3D & 38.4             & 61.3             & 92.0             & 96.6             & 88.2             & 89.3             \\
                                                            & 2s HetGCN \cite{gao2023learning}               & TNNLS 2024  & 3D           & 38.4             & 61.2             & \textbf{93.1}    & \textbf{97.2}    & 88.9             & 90.3             \\
                                                            & 4s TranSkeleton \cite{liu2023transkeleton}     & TCSVT 2023  & Transformer  & --               & --               & 92.8             & 97.0             & 89.4             & 90.5             \\
                                                            & 4s THTFormer \cite{ma2025thtformer}            & PR 2026     & Transformer  & --               & --               & 93.0             & 96.8             & \textbf{90.1}    & 91.3             \\
            \cline{2-10}
                                                            & 2s NeuroPath (Ours)                            & --          & (2+1)D       & 38.9             & 61.9             & 92.4             & 97.0             & 89.5             & 91.0             \\
                                                            & 4s NeuroPath (Ours)                            & --          & (2+1)D       & \textbf{40.0}    & \textbf{62.9}    & 93.0             & \textbf{97.2}    & 89.9             & \textbf{91.5}    \\
            \bottomrule
        \end{tabular}
    }
    \vspace{1mm}
    \begin{minipage}{0.95\linewidth}
        \footnotesize
        \textit{Note:} MotionBERT (finetune)~\cite{zhu2023motionbert} uses additional large-scale pretraining data (e.g., Human3.6M) before fine-tuning, while other methods follow standard NTU training protocols.
    \end{minipage}
\end{table*}

\subsection{Comparisons with the State-of-the-Art}

\myPara{Baselines}
We compare NeuroPath with representative state-of-the-art skeleton-based action recognition methods, including STGCN-based models~\cite{liu2020disentangling,shi2020skeleton,gao2023learning,chen2021channel,chi2022infogcn}, Transformer-based models~\cite{liu2023transkeleton,ma2025thtformer}, and recent motion representation learning methods~\cite{zhu2023motionbert}.We include MotionBERT~\cite{zhu2023motionbert} in \cref{tab-compare} and directly compare it with our method under the same evaluation protocols. MotionBERT achieves strong performance due to its large-scale motion pretraining. Therefore, to provide a fair and comprehensive comparison, we report both its training-from-scratch and fine-tuned results. For other recent approaches \cite{gao2023learning,duan2022revisiting} that are not included in the main quantitative comparison, the primary reason is the inconsistency in input representation and evaluation pipeline with the standard coordinate-based ST-GCN setting \cite{chen2021channel,shi2020skeleton} adopted in this work. These methods often rely on alternative pose representations, training pipelines, or preprocessing strategies, which makes direct comparison under identical experimental protocols non-trivial.
For example, PoseConv3D~\cite{duan2022revisiting} operates on heatmap-based pose representations using a 3D CNN architecture, which differs from coordinate-based skeleton inputs and graph convolution modeling. We thus treat it as a different pose representation paradigm rather than a directly comparable ST-GCN baseline.

\myPara{Accuracy}
\cref{tab-compare} reports the accuracy comparison on multiple benchmarks.
Here, ``Js'', ``2s'', and ``4s'' denote single-stream joint input, two-stream joint+bone inputs, and four-stream joint, bone, joint motion, and bone motion inputs, respectively.
Overall, NeuroPath is primarily designed as a single-stream GCN-based framework that improves representation learning within each individual stream. The proposed design can also be extended to multi-stream settings as a plug-in module for evaluation.
First, in the single-stream setting (Js), NeuroPath consistently achieves state-of-the-art performance among coordinate-based ST-GCN methods, demonstrating the effectiveness of the proposed intra-graph modeling. Importantly, the observed trends are consistent with \cref{fig:motivation}, where NeuroPath improves modality utilization and reduces imbalance across skeletal representations.%
Compared with MotionBERT, NeuroPath outperforms its training-from-scratch version on NTU RGB+D 60, while the 4s NeuroPath achieves the same reported accuracy as the finetuned MotionBERT on NTU RGB+D 60, i.e., 93.0\% on X-Sub and 97.2\% on X-View, without relying on large-scale motion pretraining. Second, extending to multi-stream configurations (2s and 4s) further improves performance in most cases, especially on Kinetics Skeleton 400 and NTU RGB+D 120. However, we observe that the benefit of multi-stream fusion is not always proportional, as naive aggregation of multiple streams may introduce redundancy or suboptimal cross-stream interactions.
This suggests that while our method enhances per-stream representation quality, multi-stream performance also depends on the effectiveness of cross-stream fusion strategies, which is orthogonal to our main contribution. Finally, compared with DualHead~\cite{chen2021learning}, NeuroPath achieves clear improvements on NTU RGB+D 120, validating the effectiveness of the proposed single-stream enhancement mechanism.

\myPara{Efficiency}
The computational efficiency of NeuroPath under the X-Sub protocol of NTU RGB+D 60 is illustrated in \cref{fig-bubble}, where the horizontal axis denotes GFLOPs, the vertical axis denotes accuracy, and the bubble size represents the number of parameters.
As shown in \cref{fig-bubble}, NeuroPath achieves a favorable accuracy--efficiency trade-off, attaining high recognition accuracy with relatively low computational cost and compact model size.
For the joint-stream setting, NeuroPath requires only 2.8M parameters and 4.17 GFLOPs, which is substantially more efficient than MS-G3D (3.2M parameters, 24.44 GFLOPs) and DualHead (3.0M parameters).
This efficiency advantage stems from our unified spatial–temporal modeling within each pathway, which avoids the heavy two-stage spatial–temporal convolution commonly used in STGCN variants such as CTR-GCN~\cite{chen2021channel}.
Although 3D convolution-based approaches~\cite{liu2020disentangling,gao2023learning} attempt to address this limitation, they typically incur significantly higher computational and parameter costs.
Overall, NeuroPath demonstrates that explicitly coordinated dual-pathway modeling can achieve strong performance without sacrificing efficiency.

\subsection{Ablation Study}
To investigate the contribution of each component of NeuroPath, we conducted an ablation study using the Js NeuroPath on NTU RGB+D 60 in the X-Sub setting.

\myPara{Effectiveness of Group Aggregation}
\cref{tab-group} reports the results of the proposed group aggregation strategies.
We evaluate two temporal grouping schemes: an \emph{overlapping} strategy, where adjacent groups share temporal frames, and a \emph{non-overlapping} strategy, where groups are disjoint in time.
Compared with the baseline without group aggregation, both strategies bring clear performance gains (2.1\% and 1.5\%, respectively), demonstrating the effectiveness of group-level temporal modeling.
The overlapping strategy consistently outperforms the non-overlapping one, as shared frames preserve temporal continuity across groups and enable the model to capture motion correspondence between consecutive segments, whereas non-overlapping grouping processes each segment independently and thus miss such cross-group dependencies.

\begin{table}[h]
    \centering
    \small
    \MStabcaption{tab-group}{Accuracy (\%) with various grouping strategies. The changes are relative to the first row.}
    \begin{tabular}{lll}
        \toprule
        \textbf{Configuration}      & \textbf{Param. (M)}     & \textbf{Acc. (\%)}              \\
        \midrule
        Baseline w/o Group          & 2.51                    & 88.7                            \\
        Baseline w/ Overlapping     & 2.80 $^{\uparrow 0.29}$ & \textbf{90.8} $^{\uparrow 2.1}$ \\
        Baseline w/ Non-Overlapping & 2.80 $^{\uparrow 0.29}$ & 90.2 $^{\uparrow 1.5}$          \\
        \bottomrule
    \end{tabular}
\end{table}
We further investigate the impact of different overlapping strategies on the performance of NeuroPath, as summarized in \cref{tab-window}.
Here, ``B1'', ``B2'', and ``B3'' denote the three STGGCBs in the network, and ``\ding{55}'' indicates that the corresponding block is not grouped.
The results show that grouping the first two blocks yields the best performance, while extending grouping to all three blocks introduces an additional 0.85M parameters compared to grouping only the first two blocks, increasing model complexity and leading to overfitting and degraded accuracy.
We also observe that the temporal window size plays a critical role: both excessively small and overly large windows hurt performance.
For instance, a window size of 3 results in a 0.4\% accuracy drop compared to a window size of 7.
Accordingly, we set the window size to 7 for all experiments, as it provides a better balance between temporal receptive field and sensitivity to local motion cues, i.e., larger windows tend to smooth out local dynamics, while smaller windows lack sufficient temporal context.

\begin{table}[h]
    \centering
    \small
    \caption{Accuracy (\%) with various group configurations. The changes are relative to the first row.}
    \label{tab-window}
    \begin{tabular}{ccccll}
        \toprule
        \multirow{2}{*}{\textbf{Configuration}}  &
        \multicolumn{3}{c}{\textbf{Window Size}} & \multirow{2}{*}{\textbf{Param. (M)}} & \multirow{2}{*}{\textbf{Acc. (\%)}}                                                                           \\\cline{2-4}
                                                 & B1                                   & B2                                  & B3        &                          &                                  \\
        \midrule
        w/o Group                                & \ding{55}                            & \ding{55}                           & \ding{55} & 2.51                     & 88.7                             \\ \hline
        \multirow{7}{*}{w/ Group}                & \ding{51}                            & \ding{55}                           & \ding{55} & 2.65 $^{\uparrow 0.14}$  & 90.4  $^{\uparrow 1.7}$          \\
                                                 & \ding{51}                            & \ding{51}                           & \ding{55} & 2.80 $^{\uparrow 0.29}$  & \textbf{90.8}  $^{\uparrow 2.1}$ \\
                                                 & \ding{51}                            & \ding{51}                           & \ding{51} & 3.36  $^{\uparrow 0.85}$ & 90.4  $^{\uparrow 1.7}$          \\
                                                 & \ding{55}                            & \ding{51}                           & \ding{55} & 2.66  $^{\uparrow 0.15}$ & 90.1  $^{\uparrow 1.4}$          \\
                                                 & \ding{55}                            & \ding{55}                           & \ding{51} & 3.07  $^{\uparrow 0.56}$ & 89.9  $^{\uparrow 1.2}$          \\
                                                 & \ding{51}                            & \ding{55}                           & \ding{51} & 3.21  $^{\uparrow 0.70}$ & 90.3  $^{\uparrow 1.6}$          \\
                                                 & \ding{55}                            & \ding{51}                           & \ding{51} & 3.22  $^{\uparrow 0.71}$ & 90.2  $^{\uparrow 1.5}$          \\
        \bottomrule
    \end{tabular}
\end{table}

Additionally, \cref{tab-plug} reports the plug-and-play results of integrating the proposed modules into AGCN~\cite{shi2020skeleton}.
Introducing the attention mask $\mathbf{M}_{\mathrm{att}}$ yields a moderate accuracy improvement with a small parameter increase, while incorporating the proposed group aggregation module leads to a substantially larger performance gain.
Combining both components further improves accuracy, demonstrating their complementary effects.
These results indicate that the proposed group aggregation module is not only compatible with existing STGCN architectures, but also consistently improves their representation capability under a modest parameter overhead.

\begin{table}[h]
    \centering
    \caption{Plug-and-play accuracy (\%) of proposed modules. The changes are relative to the first row.}
    \label{tab-plug}

    \begin{tabular}{llll}
        \toprule
        \small
        \textbf{Configuration}                           & \textbf{Param. (M)}     & \textbf{Acc. (\%)}     \\
        \midrule
        AGCN (Js) \cite{shi2020skeleton}                 & 3.47                    & 87.4                   \\
        AGCN (Js) w/ $\mathbf{M}_{\mathrm{att}}$         & 3.87 $^{\uparrow 0.40}$ & 88.2 $^{\uparrow 0.8}$ \\
        AGCN (Js) w/ Group                               & 4.45 $^{\uparrow 0.98}$ & 89.1 $^{\uparrow 1.7}$ \\
        AGCN (Js) w/ Group + $\mathbf{M}_{\mathrm{att}}$ & 4.85 $^{\uparrow 1.38}$ & 89.7 $^{\uparrow 2.3}$ \\
        \bottomrule
    \end{tabular}
\end{table}

\myPara{Effectiveness of Adaptive Structural Aggregation}
\cref{tab-arch} shows the results of various structural aggregation strategies, among which the element-wise product ($\odot$) outperforms the additive strategy ($+$) proposed in the previous work \cite{liu2020disentangling} by 0.6\%.
This demonstrates the effectiveness of our adaptive structural aggregation.
In addition, \cref{tab-setting} presents the results of different multi-hop aggregation strategies, where `T', `M', and `B' correspond to the top, middle, and bottom aggregation strategies in \cref{fig-decomposition}.
NeuroPath achieves the best performance among them, highlighting its effectiveness.
By combining the results from \cref{tab-arch,tab-setting}, it confirms that the importance-unaware aggregation strategy (90.6\%) is inflexible, and adding residual links (90.2\%) introduces redundancy, thereby limiting action recognition performance.

\begin{table}[!h]
    \centering
    \caption{
        Accuracy (\%) in various spatial and temporal configurations. The changes are relative to the third row.}
    \label{tab-setting}
    \begin{tabular}{cccccll}
        \toprule
        \multicolumn{5}{c}{\textbf{Configuration}} & \multirow{2}{*}{\textbf{Param. (M)}} & \multirow{2}{*}{\textbf{Acc. (\%)}}                                                                               \\\cline{1-5}
        T                                          & M                                    & B                                   & MK        & MD        &                         &                           \\
        \midrule
        \ding{51}                                  & \ding{55}                            & \ding{55}                           & \ding{55} & \ding{51} & 2.80                    & 90.2                      \\
        \ding{55}                                  & \ding{51}                            & \ding{55}                           & \ding{55} & \ding{51} & 2.80                    & 90.6                      \\
        \ding{55}                                  & \ding{55}                            & \ding{51}                           & \ding{55} & \ding{51} & 2.80                    & \textbf{90.8}             \\
        \ding{55}                                  & \ding{55}                            & \ding{51}                           & \ding{51} & \ding{55} & 3.04 $^{\uparrow 0.24}$ & 90.5  $^{\downarrow 0.3}$ \\
        \bottomrule
    \end{tabular}
\end{table}

\myPara{Effectiveness of Multi-Scale Temporal Convolution}
We investigate multi-kernel (MK) and multi-dilation (MD) temporal convolutions, as summarized in \cref{tab-setting}.
Compared to MK, MD achieves higher accuracy with fewer parameters, indicating superior parameter efficiency.
When combined with the proposed aggregation strategy, MD further improves performance while keeping the model compact.
These results suggest that MD provides a more effective temporal modeling mechanism and works particularly well with our spatial aggregation design, offering a favorable balance between temporal receptive field and model efficiency for action recognition.

\myPara{Effectiveness of Spatial-Temporal Attention}
As shown in the sixth row of \cref{tab-arch}, removing the spatial–temporal attention module ($\mathbf{M}_{\mathrm{att}}$) results in a 0.9\% accuracy drop, indicating its clear contribution to performance.
In addition, the plug-and-play results in \cref{tab-plug} show that incorporating $\mathbf{M}_{\mathrm{att}}$ into AGCN leads to a consistent 0.8\% accuracy improvement.
Together, these results demonstrate that spatial–temporal attention provides effective adaptive weighting of joint interactions over space and time, contributing to more discriminative representations for action recognition.

\begin{table}[!h]
    \centering
    \small
    \caption{Accuracy (\%) with various configurations. The changes are relative to the first row.}
    \label{tab-arch}
    \begin{tabular}{lll}
        \toprule
        \textbf{Configuration}                                                            & \textbf{Param. (M)}       & \textbf{Acc. (\%)}       \\
        \midrule
        \cref{fig-architecture-a} w/ $\odot  \mathbf{M}_{k}$ (see \cref{eq_mk})           & 2.80                      & \textbf{90.8}            \\
        \cref{fig-architecture-a} w/ $+ \mathbf{M}_{k}$ (see \cref{eq_mk})                & 2.80 $^{\uparrow 0.00}$   & 90.2 $^{\downarrow 0.6}$ \\
        \midrule
        \cref{fig-architecture-b} w/ $\odot  \mathbf{M}_{k}$ (see \cref{eq_mk})           & 1.40 $^{\downarrow 1.40}$ & 88.5 $^{\downarrow 2.3}$ \\
        \cref{fig-architecture-c} w/ $\odot  \mathbf{M}_{k}$ (see \cref{eq_mk})           & 2.80 $^{\uparrow 0.00}$   & 90.5 $^{\downarrow 0.3}$ \\
        \midrule
        \cref{fig-architecture-a} w/o $\mathcal{M}_{\mathrm{s}},\mathcal{M}_{\mathrm{t}}$ & 2.80 $^{\uparrow 0.00}$   & 89.7 $^{\downarrow 1.1}$ \\
        \cref{fig-architecture-a} w/o $\mathbf{M}_{\mathrm{att}}$ (see \cref{eq_att})     & 2.23 $^{\downarrow 0.57}$ & 89.9 $^{\downarrow 0.9}$ \\ \midrule
        \cref{fig-architecture-a} w/o Dynamic Fusion                                      & 2.80 $^{\uparrow 0.00}$   & 90.4 $^{\downarrow 0.4}$ \\
        \cref{fig-architecture-a} w/o $\alpha_1,\beta_1$                                  & 2.80 $^{\uparrow 0.00}$   & 90.3 $^{\downarrow 0.5}$ \\
        \cref{fig-architecture-a} w/o $\alpha_2,\beta_2$                                  & 2.80 $^{\uparrow 0.00}$   & 90.5 $^{\downarrow 0.3}$ \\
        \cref{fig-architecture-a} w/o $\alpha_3,\beta_3$                                  & 2.80 $^{\uparrow 0.00}$   & 90.4 $^{\downarrow 0.4}$ \\
        \bottomrule
    \end{tabular}
\end{table}

\myPara{Effectiveness of Dual-pathway Architecture}
The results of different architectural designs are reported in the first four rows of \cref{tab-arch}.
Both dual-pathway variants (see \cref{fig-architecture-a} and \cref{fig-architecture-c}) outperform the single-pathway baseline (see \cref{fig-architecture-b}), demonstrating the effectiveness of separating spatial-aware and temporal-aware processing.
Moreover, our proposed architecture (see \cref{fig-architecture-a}) achieves a 0.3\% accuracy improvement over (c), indicating that dynamic fusion enables more effective interaction between the two pathways than hard fusion.

\myPara{Effectiveness of Modality Transformation}
We further evaluate the contribution of the modality transformation modules $\mathcal{M}_\mathrm{s}$ and $\mathcal{M}_\mathrm{t}$ by removing them in \cref{tab-arch}.
Without these modules, both pathways receive identical inputs, which leads to a 1.1\% accuracy drop.
This result highlights the importance of explicitly assigning complementary modalities to the spatial-aware and temporal-aware pathways.
In addition, compared with prior work that processes a single modality in a single-stream network~\cite{Wang_2023_CVPR}, where the bone-motion stream achieves 88.0\% accuracy, our method reaches 90.7\%.
This comparison further demonstrates the benefit of jointly leveraging complementary spatial-aware and temporal-aware information for action recognition.

\myPara{Effectiveness of Dynamic Fusion}
As shown in \cref{tab-arch}, removing the dynamic fusion module results in a 0.4\% accuracy drop, confirming its positive contribution.
We further analyze the effect of fusion at different network stages by individually removing the fusion weights in each block, which leads to varying degrees of performance degradation.
These results indicate that dynamic fusion consistently contributes across multiple stages of the network.
While most state-of-the-art methods in \cref{tab-compare} rely on late fusion strategies, earlier fusion approaches~\cite{duan2022revisiting} often introduce additional complexity or require richer input modalities.
In contrast, our dynamic fusion mechanism enables effective stage-wise interaction within a single model, providing a more balanced trade-off between performance and model complexity.

\myPara{Effectiveness of Multi-Stream Skeletal Modeling}
To evaluate the effectiveness of multi-stream modeling and the benefit of jointly using joint- and bone-based features, we conduct an ablation study on different skeletal modality combinations, as shown in \cref{fig:empirical-ours}.
The evaluated settings include single-modality inputs, all pairwise combinations, and the full four-stream (4s) configuration.
Single-modality inputs achieve comparable but limited performance (e.g., 90.82\% for joint and 90.85\% for bone), whereas combining joint and bone features yields a clear improvement to 92.40\%, demonstrating their complementarity.
Further gains are consistently obtained when incorporating motion cues, and the full 4s setting achieves the best performance (93.0\%).
These results confirm that multi-stream modeling effectively exploits complementary structural and motion information, leading to more robust representations.

\subsection{Qualitative and Quantitative Results}
We show more qualitative and quantitative results.
All results are based on Js NeuroPath in the X-Sub setting of NTU RGB+D 60.

\begin{figure}
    \centering
    \includegraphics[width=\linewidth]{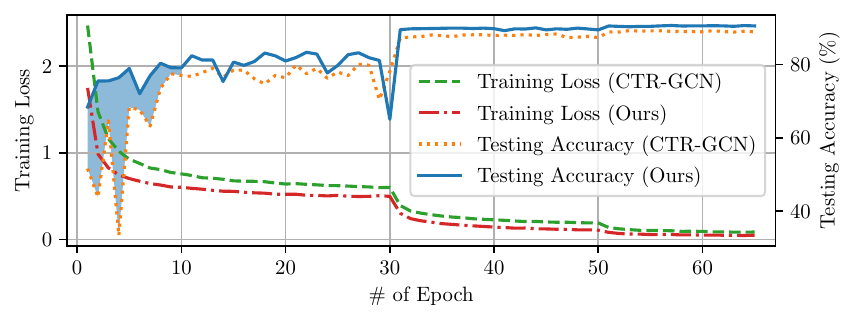}
    \caption{
        A comparison of the training loss and testing accuracy between CTR-GCN \cite{chen2021channel} and our method.
        The results demonstrate the effectiveness of our proposed model, as it achieves lower training loss and higher testing accuracy compared to CTR-GCN.
    }
    \label{fig-loss_acc}
\end{figure}

\myPara{Fast Convergence}
In \cref{fig-loss_acc}, the training loss and testing accuracy curves for NeuroPath and CTR-GCN \cite{chen2021channel} are presented. The blue shading in the first 10 epochs represents the training phase. Our method demonstrates faster convergence than CTR-GCN, and around the 30th epoch, the learning rate changes noticeably. The results show that NeuroPath achieves lower training loss and higher testing accuracy than CTR-GCN, indicating its efficacy in modeling spatial-temporal features for action recognition. Furthermore, NeuroPath converges faster in the early stages, delivering superior performance with fewer training epochs. This shows the superiority of NeuroPath.

\begin{figure}
    \centering
    \subfigure{\label{fig-fusion_confusion_a}
        \begin{overpic}[height=0.45\linewidth]{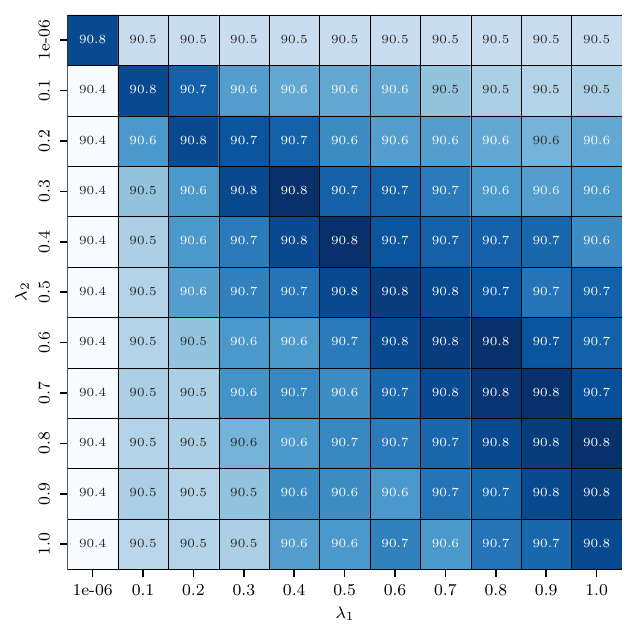}
            \put(1,2){\footnotesize(a)}
        \end{overpic}
    }\subfigure{\label{fig-fusion_confusion_b}
        \begin{overpic}[height=0.45\linewidth]{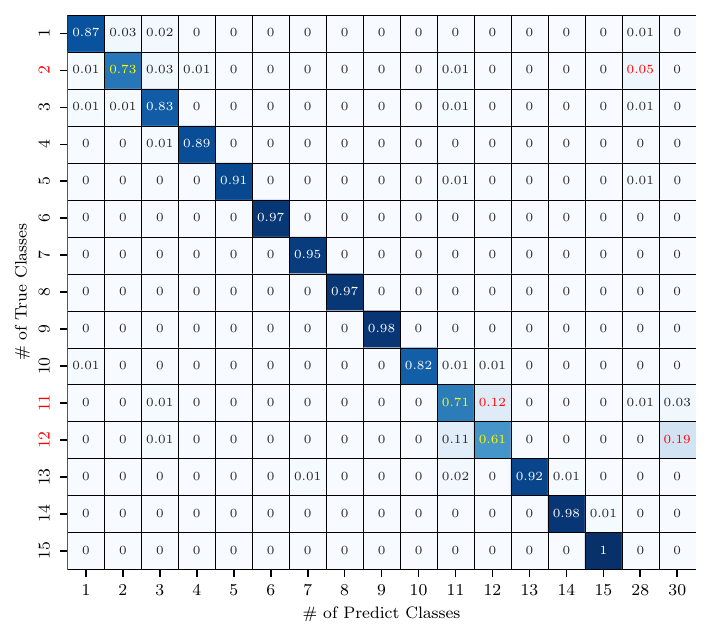}
            \put(1,2){\footnotesize(b)}
        \end{overpic}
    }
    \caption{
        Heatmaps of (a) the fusion matrix and (b) the confusion matrix.
        $\lambda_1$ and $\lambda_2$ are coefficient weights.
    }
    \label{fig-fusion_confusion}
\end{figure}

\myPara{Effective Fusion}
\cref{fig-fusion_confusion_a} visualizes the confusion matrices under different fusion settings.
We observe that using either spatial or temporal pathway alone yields inferior performance compared to their fused counterparts, indicating that neither pathway is sufficient in isolation.
In contrast, dual-pathway fusion consistently improves recognition accuracy, validating the benefit of jointly modeling spatial structure and temporal dynamics.
In all experiments, the fusion coefficients are set to $\lambda_1=\lambda_2=1$.
Notably, the performance gains are robust across different fusion configurations, suggesting that NeuroPath does not rely on a specific weighting scheme.
This robustness stems from the dynamic fusion mechanism, which enables progressive and bidirectional information exchange between the two pathways while preserving pathway-specific representations.
These results demonstrate that effective coordination between spatial-aware and temporal-aware pathways, rather than the dominance of a single pathway, is the key factor underlying the performance improvements of NeuroPath.

\begin{figure}
    \centering
    \subfigure[Writing (311-th sample) $\rightarrow$ {typing on a keyboard}]{
        \includegraphics[width=0.15\linewidth]{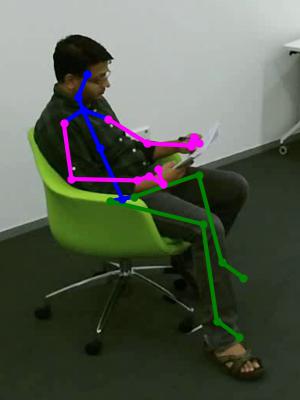}
        \includegraphics[width=0.15\linewidth]{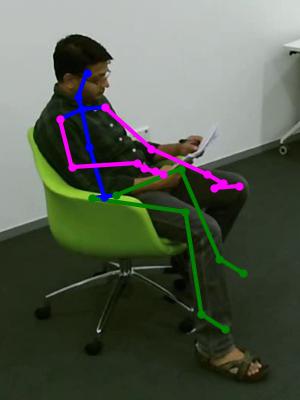}
        \includegraphics[width=0.15\linewidth]{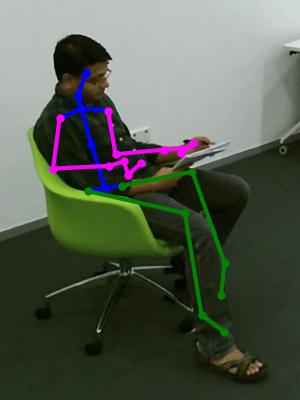}
        \includegraphics[width=0.15\linewidth]{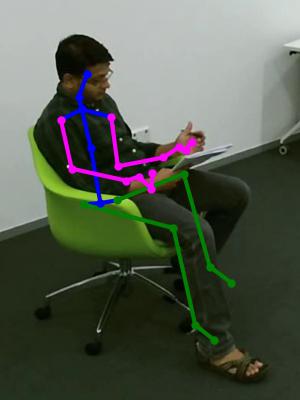}
        \includegraphics[width=0.15\linewidth]{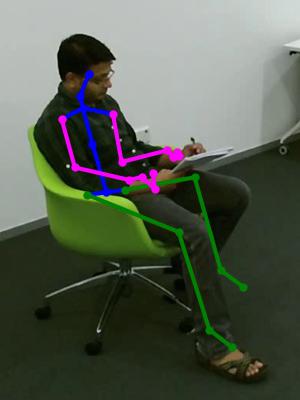}
        \includegraphics[width=0.15\linewidth]{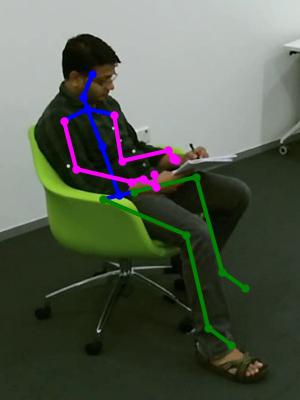}
        \label{fig-failure_a}
    }\\
    \subfigure[Typing on a keyboard (3326-th sample) $\rightarrow$ {writing}]{
        \includegraphics[width=0.15\linewidth]{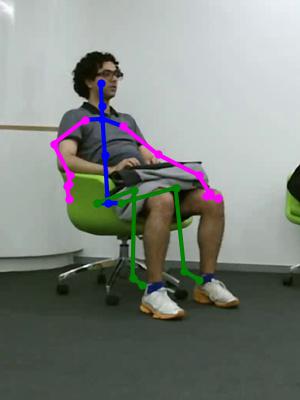}
        \includegraphics[width=0.15\linewidth]{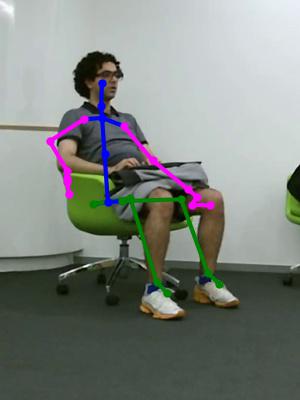}
        \includegraphics[width=0.15\linewidth]{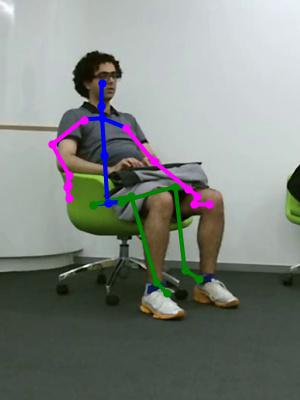}
        \includegraphics[width=0.15\linewidth]{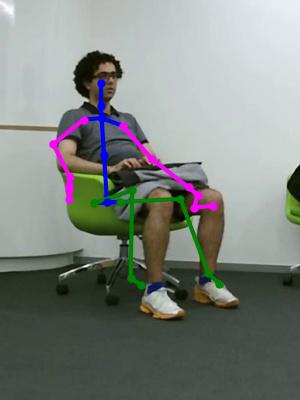}
        \includegraphics[width=0.15\linewidth]{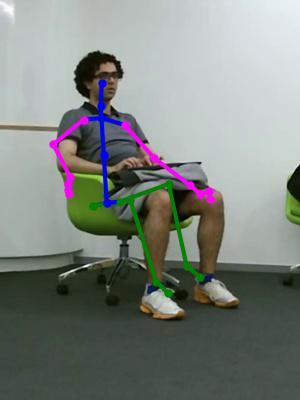}
        \includegraphics[width=0.15\linewidth]{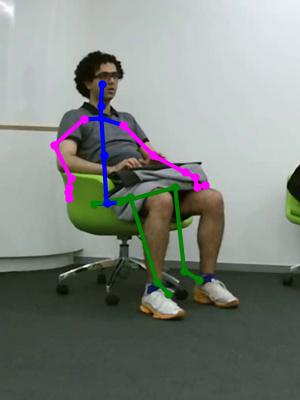}

        \label{fig-failure_b}
    }
    \caption{
        Selected frames with their human skeletons of two failure cases.
    }
    \label{fig-failure}
\end{figure}

\myPara{Failure Cases}
The partial inter-class confusion matrix shown in \cref{fig-fusion_confusion_b} reveals that some actions share similarities in terms of body movements, leading to high confusion rates between them.
For example, the 12-th \textit{writing}, has been confused with the 30-th \textit{typing on a keyboard}, in 19\% of the samples.
We provide visual examples of these two confused actions in \cref{fig-failure}, where different limbs are highlighted with different colors.
In \cref{fig-failure_a}, the 311-th sample of \textit{writing} is misclassified as \textit{typing on a keyboard}, while in \cref{fig-failure_b}, the 3326-th sample of \textit{typing on a keyboard} is misclassified as \textit{writing}.
However, the hand and forearm movements are too subtle for NeuroPath to recognize.
Our findings underscore the challenge of recognizing similar actions and the importance of developing robust features to capture subtle differences between them.
It is noted that NeuroPath outperforms MS-G3D \cite{liu2020disentangling} in recognizing  \textit{writing}, achieving an accuracy of 61\% compared to that of 54\%. Furthermore, NeuroPath exhibits a lower misclassification rate of \textit{writing} as \textit{typing on a keyboard} at 19\%, whereas MS-G3D has a higher rate of 23\%.
These results further demonstrate the superiority of NeuroPath over existing state-of-the-art approaches.

\begin{figure}
    \centering
    \begin{overpic}[width=\linewidth, trim=0 35 0 10, clip]{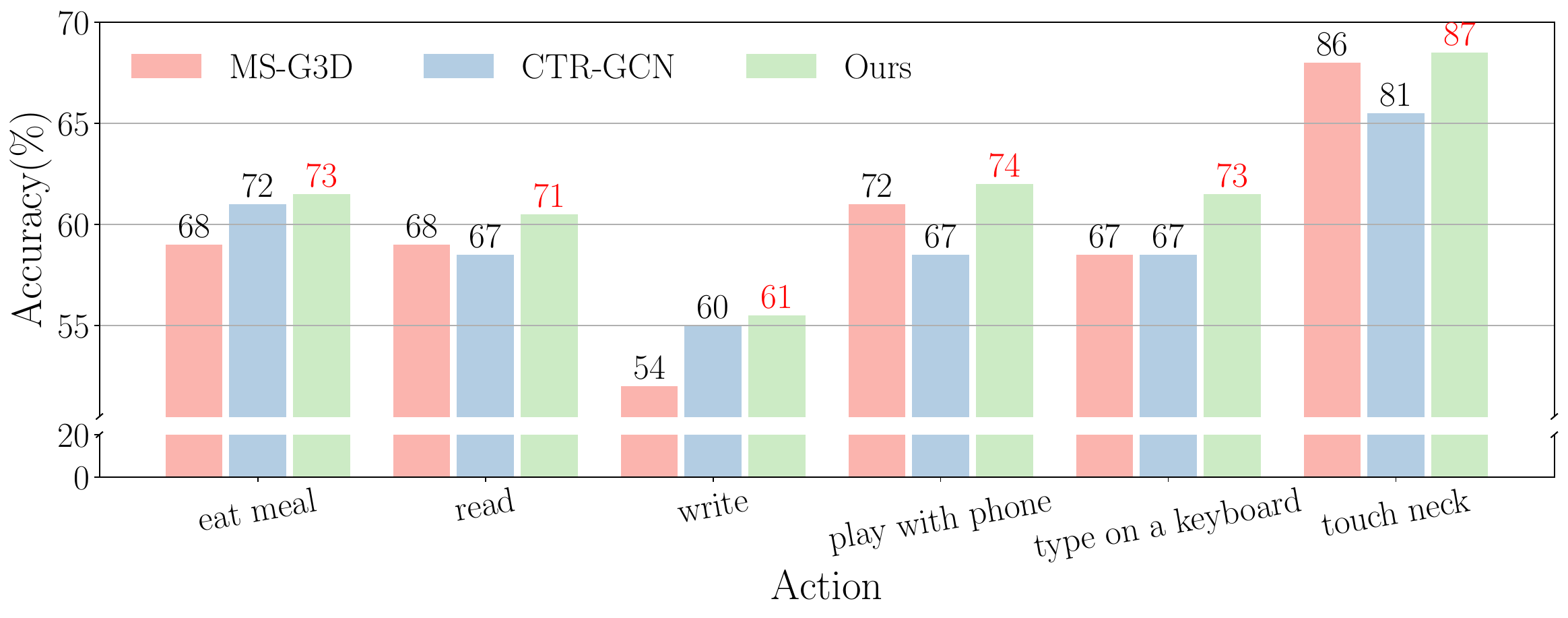}
    \end{overpic}
    \caption{
        An accuracy comparison of six challenging classes in the X-Sub setting on NTU RGB+D 60.
    }
    \label{fig-challenging}
\end{figure}

\begin{figure}
    \centering
    \subfigure[Touching neck]{
        \includegraphics[width=0.15\linewidth]{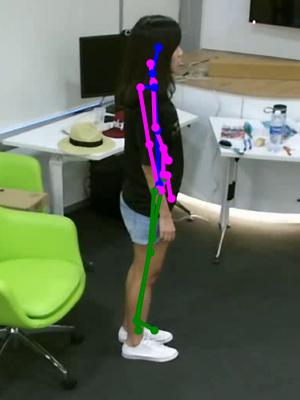}
        \includegraphics[width=0.15\linewidth]{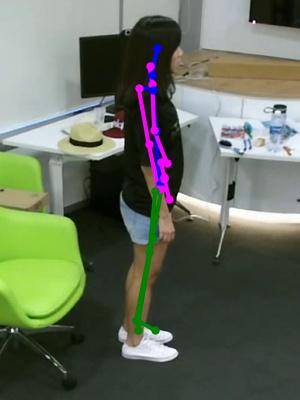}
        \includegraphics[width=0.15\linewidth]{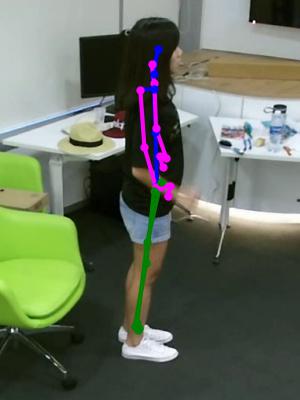}
        \includegraphics[width=0.15\linewidth]{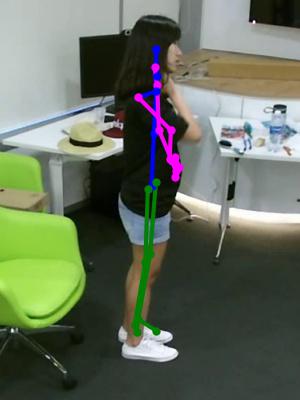}
        \includegraphics[width=0.15\linewidth]{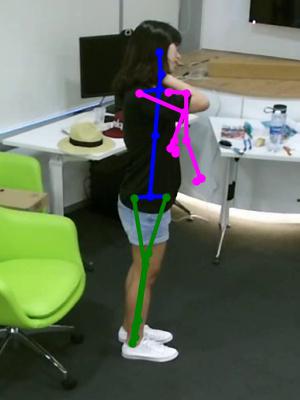}
        \includegraphics[width=0.15\linewidth]{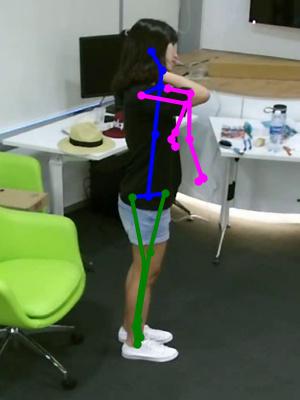}
        \label{fig-challenging-1_touch}
    }\\
    \subfigure[Brushing hair]{\label{fig-brushing}
        \includegraphics[width=0.15\linewidth]{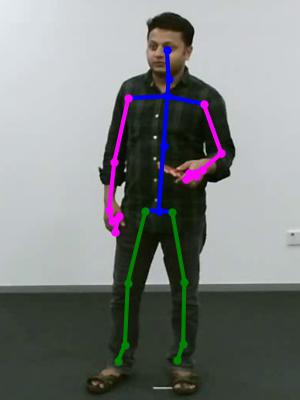}
        \includegraphics[width=0.15\linewidth]{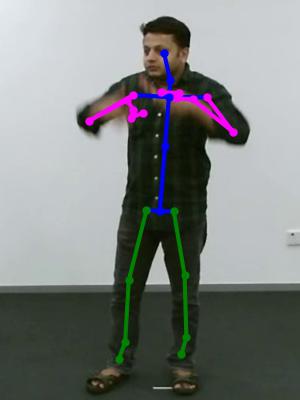}
        \includegraphics[width=0.15\linewidth]{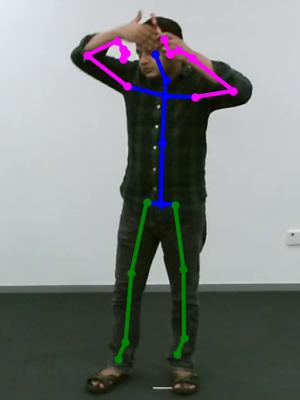}
        \includegraphics[width=0.15\linewidth]{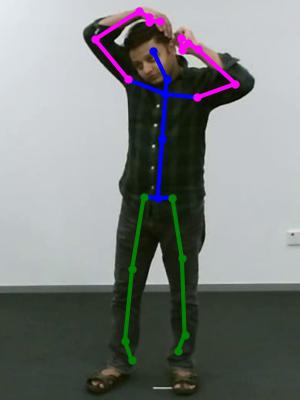}
        \includegraphics[width=0.15\linewidth]{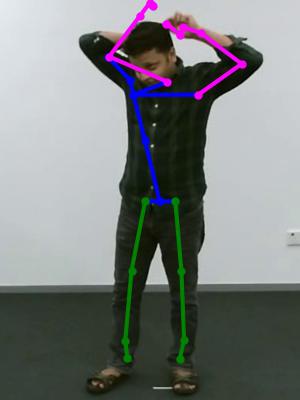}
        \includegraphics[width=0.15\linewidth]{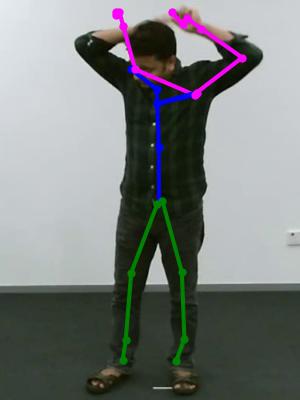}
    }
    \caption{Selected challenging and well-recognized samples.
    }
    \label{fig-challenging-1}
\end{figure}

\myPara{Effectiveness of Long-Range Dependencies}
To provide further insights into the effectiveness of NeuroPath in recognizing challenging actions, we have selected six challenging actions in \cref{fig-challenging-1}, and the result is shown in \cref{fig-challenging}.
Additionally, \cref{tab-challenging} reports the predicted scores of CTR-GCN \cite{chen2021channel} and our NeuroPath.
Ours outperforms CTR-GCN in all the actions, demonstrating the effectiveness of incorporating long-range dependency information in NeuroPath.
For instance, \textit{touching neck} in \cref{fig-challenging-1_touch} and \textit{brushing hair} in \cref{fig-brushing} are hard to distinguish as they both involve hand movement close to the head. However, the subtle difference is mainly in the interaction of different parts of the human body rather than the local hand movements. NeuroPath achieves a correct recognition rate of 98\%, demonstrating the effectiveness of incorporating long-range dependencies in distinguishing the two actions.

\begin{table}[!h]
    \centering
    \caption{Prediction scores of CTR-GCN \cite{chen2021channel} and NeuroPath. }
    \label{tab-challenging}
    \resizebox{0.8\linewidth}{!}{
        \begin{tabular}{cccc}
            \toprule
            \#                 & \textbf{Action}                                 & \textbf{CTR-GCN \cite{chen2021channel}} & \textbf{Ours} \\
            \midrule
            \multirow{2}{*}{1} & \cellcolor{LightCyan} Reading (GT)              & 0.21                                    & 0.82          \\
                               & \cellcolor{LightRed} Typing on a keyboard       & 0.53                                    & 0.05          \\
            \midrule
            \multirow{2}{*}{2} & \cellcolor{LightCyan} Typing on a keyboard (GT) & 0.12                                    & 0.59          \\
                               & \cellcolor{LightRed} Writing                    & 0.47                                    & 0.28          \\
            \midrule
            \multirow{2}{*}{3} & \cellcolor{LightCyan} Touching neck (GT)        & 0.02                                    & 0.88          \\
                               & \cellcolor{LightRed} Brushing hair              & 0.96                                    & 0.10          \\
            \bottomrule
        \end{tabular}
    }
\end{table}

\begin{figure}
    \centering

    \subfigure[Spatial-aware pathway]{\label{fig-heatmap_a}
        \begin{overpic}[height=0.21\linewidth]{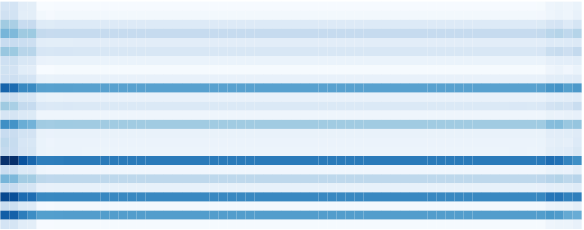}
            \put(0,0){\tikz\fill[white,fill opacity=0.75](0,0)rectangle(0.3,0.3);}
            \put(0,4.5){\linethickness{0.25mm}\color{red}\polygon(0,0)(100,0)(100,2)(0,2)}\put(0,1.5){\linethickness{0.25mm}\color{red}\polygon(0,0)(100,0)(100,2)(0,2)}\put(1,2){\scriptsize\color{black}s1}
            \multiput(26,20)(0.5, -2){7}{\linethickness{0.1mm}\color{blue}\line(1, -4){0.2}}
            \put(15,20){\scriptsize Left hand}
            \multiput(66,20)(-0.5, -2){9}{\linethickness{0.1mm}\color{blue}\line(-1, -4){0.2}}
            \put(55,20){\scriptsize Right hand}
        \end{overpic}
        \begin{overpic}[height=0.21\linewidth]{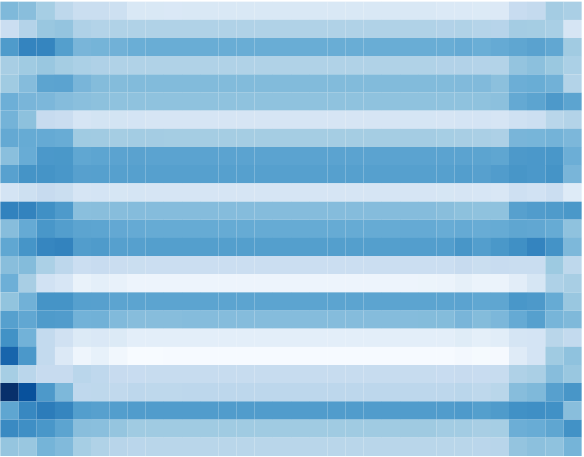}
            \put(0,0){\tikz\fill[white,fill opacity=0.75](0,0)rectangle(0.3,0.3);}
            \put(1,2){\scriptsize\color{black}s2}
        \end{overpic}
        \begin{overpic}[height=0.21\linewidth]{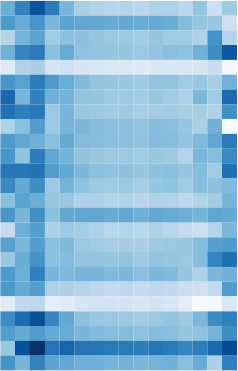}
            \put(0,0){\tikz\fill[white,fill opacity=0.75](0,0)rectangle(0.3,0.3);}
            \put(1,2){\scriptsize\color{black}s3}
        \end{overpic}}

    \subfigure[Temporal-aware pathway]{\label{fig-heatmap_b}
        \begin{overpic}[height=0.21\linewidth]{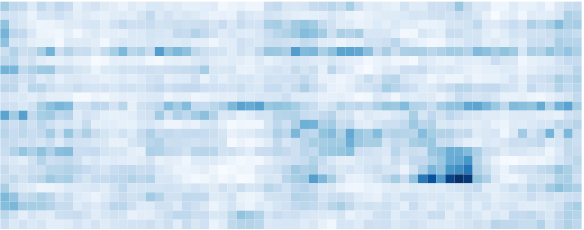}
            \put(0,0){\tikz\fill[white,fill opacity=0.75](0,0)rectangle(0.3,0.3);}
            \put(1,2){\scriptsize\color{black}t1}
        \end{overpic}
        \begin{overpic}[height=0.21\linewidth]{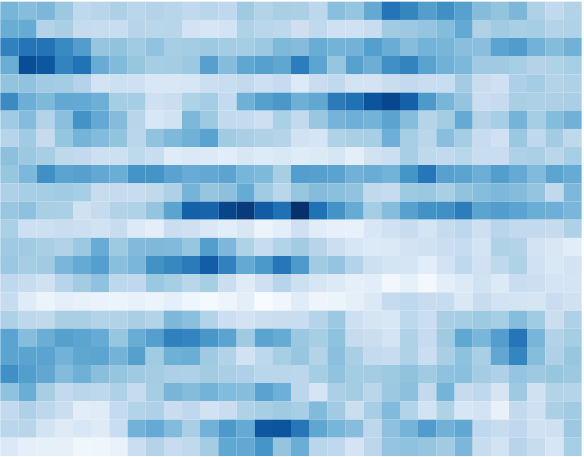}
            \put(0,0){\tikz\fill[white,fill opacity=0.75](0,0)rectangle(0.3,0.3);}
            \put(1,2){\scriptsize\color{black}t2}
        \end{overpic}
        \begin{overpic}[height=0.21\linewidth]{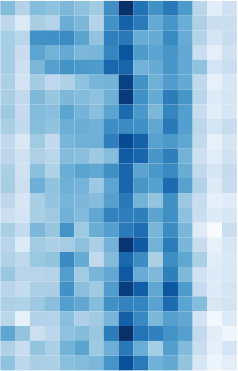}
            \put(0,0){\tikz\fill[white,fill opacity=0.75](0,0)rectangle(0.3,0.3);}
            \put(1,2){\scriptsize\color{black}t3}
        \end{overpic}}

    \caption{
        Attention maps in the first STGGCB: (a) `s1', `s2', and `s3' are heatmaps of three STGGCBs in the spatial-aware pathway, (b) `t1', `t2', and `t3'  are heatmaps of three STGGCBs in the temporal-aware pathway. The darker color denotes the higher attention weight for the corresponding joint.
    }
    \label{fig-attention}
\end{figure}

\myPara{Attention Maps}
\cref{fig-attention} shows attention maps, specifically for a sample in \cref{fig-brushing} of \textit{brushing hair}. The vertical axis denotes the joint index while the horizontal axis denotes the frame index. Darker colors indicate active joints with more discriminative features.
Notably, the hand joints exhibit stronger attention in \cref{fig-heatmap_a}, ensuring the preservation of crucial information.
Additionally, \cref{fig-heatmap_a} attends to more variations in the spatial domain across joints, while \cref{fig-heatmap_b} attends to more variations in the temporal domain across frames, indicating the presence of unique information in each pathway.
These findings emphasize NeuroPath's effectiveness in enhancing the discriminative features and improving action recognition accuracy.

\section{Conclusion}
\label{sec:conclusion}
This paper revisits skeleton-based action recognition from a modality-aware perspective.
While STGCNs have achieved remarkable progress,
we identify two persistent limitations: inefficient spatial–temporal cohesion within single branches
and inadequate exploitation of complementary skeletal modalities.
To address these issues, we propose \emph{NeuroPath}, a brain-inspired dual-pathway framework that
explicitly separates and coordinates spatial-aware and temporal-aware representations.
NeuroPath integrates a spatial–temporal group graph convolution block for coherent intra-pathway modeling,
a modality transformation unit for structured input decomposition,
and dynamic inter-pathway fusion for progressive feature interaction.
Extensive experiments on three large-scale benchmarks
demonstrate that NeuroPath consistently achieves state-of-the-art performance with improved computational efficiency.
The results validate that explicitly modeling the imbalanced yet complementary nature of skeletal modalities
leads to more effective and compact representations, reducing reliance on heavy multi-stream ensembles.

\myPara{Limitations and Future Work}
While NeuroPath focuses on improving single-stream representation learning and achieves strong performance under both single-stream and multi-stream evaluation settings, an interesting future direction is to explore more advanced multi-stream fusion strategies. In particular, modeling cross-stream interactions among joint, bone, and motion modalities in a more adaptive manner may further improve performance beyond simple aggregation. We consider such extensions to be orthogonal to the proposed single-stream enhancement framework. Beyond multi-stream fusion, another promising direction is to investigate more structured grouping and segmentation strategies, which may better capture fine-grained action semantics and further strengthen spatial–temporal coordination.
Beyond the standard skeleton-based setting, the proposed dual-pathway framework may be extended to richer skeletal representations, such as heatmap-based pose encodings, as well as to multi-modal inputs including RGB, depth, or audio cues. At a higher level, leveraging pretrained backbone models and fine-tuning them on target domains~\cite{zhu2023motionbert,zhou2026sdpt} remain a promising direction for advancing skeleton-based action understanding.
Overall, NeuroPath provides an effective framework for unified spatial–temporal modeling, and offers useful insights for developing robust and interpretable action recognition systems.

\section*{Acknowledgements}
This work was supported in part by the National Natural Science Foundation of China (No. 62272019) and the China Postdoctoral Science Foundation (No. 2025M781489).

\bibliographystyle{elsarticle-num}
\bibliography{refs}





\end{document}